\documentclass[10pt,twocolumn,letterpaper]{article}

\usepackage[pagenumbers]{cvpr} 

\usepackage{comment}

\renewcommand{\paragraph}[1]{\noindent\textbf{#1.}}

\newcommand{\modelname}{CUBE\xspace}
\newcommand{\modelnamelong}{\modelname (Control-based Unified B-spline Encoding)\xspace}

\definecolor{cvprblue}{rgb}{0.21,0.49,0.74}
\usepackage[pagebackref,breaklinks,colorlinks,allcolors=cvprblue]{hyperref}
\usepackage{booktabs} 
\usepackage{multirow} 
\usepackage{colortbl}

\def\paperID{3039} 
\def\confName{CVPR}
\def\confYear{2026}

\title{Representing 3D Faces with Learnable B-Spline Volumes}

\author{Prashanth Chandran \quad Daoye Wang \quad Timo Bolkart\\
Google\\
{\tt\small \{prchandran, daoye, tbolkart\}@google.com}
}

\begin{document}
\twocolumn[{%
   \renewcommand\twocolumn[1][]{#1}%
   \maketitle
   \begin{center}
      \centering
      \captionsetup{type=figure}
      \includegraphics[width=1.0\textwidth]{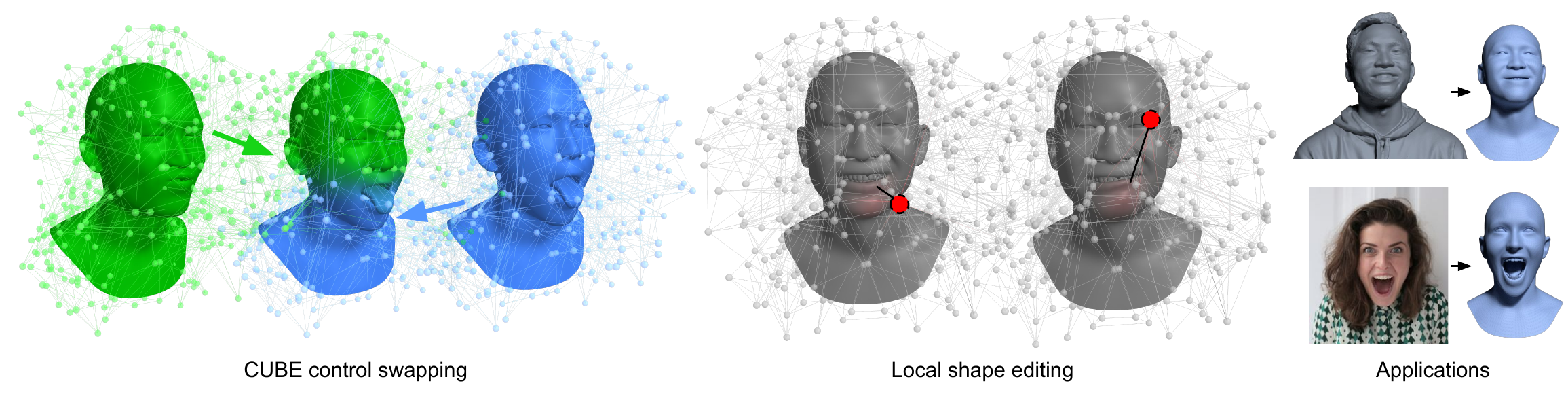}
      \caption{We present \modelnamelong, a new geometric representation for faces that extends traditional B-Spline volumes with learned features. \modelname's control features locally influence a face shape and therefore allow for precise shape editing by (left) control swapping or (middle) interactive editing. We demonstrate the usefulness of \modelname with two (right) applications; feed-forward facial scan registration where \modelname achieves state-of-the-art results and image-based regression.}
      \label{fig:teaser}
   \end{center}
   \vspace{-1mm}
}]



\newcommand{\R}[1]{\mathbb{R}^{#1}}
\newcommand{\inR}[1]{\in \R{#1}}

\newcommand{\scalar}[1]{\lowercase{#1}}
\renewcommand{\vector}[1]{\textbf{\lowercase{#1}}}
\renewcommand{\matrix}[1]{\textbf{\uppercase{#1}}}
\newcommand{\tensor}[1]{\mathcal{\uppercase{#1}}}
\newcommand{\mappingfunction}[1]{\uppercase{#1}}
\newcommand{\misc}[1]{\uppercase{#1}}

\newcommand{\gridsize}{\scalar{M}}
\newcommand{\totalgridsize}{\scalar{M}_c}

\newcommand{\griddecrementsize}{\scalar{N}}


\newcommand{\basis}[2]{\mappingfunction{N}_{#1}^{#2}}
\newcommand{\basisfunction}[3]{\basis{#1}{#2}({#3})}
\newcommand{\weight}[3]{\scalar{h}_{#1#2#3}}
\newcommand{\controlpoint}[3]{\vector{v}_{#1#2#3}}
\newcommand{\controlfeature}[3]{\vector{c}_{#1#2#3}}
\newcommand{\knots}{\scalar{t}}
\newcommand{\knotvector}{\misc{T}}
\newcommand{\degree}{\scalar{r}}
\newcommand{\featuredim}{\scalar{d}}

\newcommand{\nurbsfunction}{f}
\newcommand{\nurbsfunctionparams}[1]{\nurbsfunction(#1)}
\newcommand{\mlp}{g}
\newcommand{\mlpfunction}[1]{\mlp(#1)}

\newcommand{\featurevector}{\vector{z}}
\newcommand{\point}{\vector{x}}
\newcommand{\pointsample}{\vector{x}_{\text{sample}}}
\newcommand{\basepoint}{\vector{x}_{\text{base}}}
\newcommand{\pointrefinement}{\vector{x}_{\text{refinement}}}
\newcommand{\outvector}{\vector{x}_{\text{out}}}

\newcommand{\scanvertices}{\vector{s}}
\newcommand{\numscanvertices}{\scalar{s}}

\newcommand{\nurbs}{\uppercase{nurbs~}}

\newcommand{\secref}[1]{Sec.~\ref{#1}}
\newcommand{\figref}[1]{Fig.~\ref{#1}}
\newcommand{\tabref}[1]{Table.~\ref{#1}}

\definecolor{tabfirst}{rgb}{1, 0.7, 0.7} 
\definecolor{tabsecond}{rgb}{1, 0.85, 0.7} 
\definecolor{tabthird}{rgb}{0.5, 0.5, 0.5} 

\definecolor{red}{rgb}{1, 0.3, 0.3} 
\definecolor{orange}{rgb}{1.0, 0.6, 0.0} 
\definecolor{blue}{rgb}{0.1, 0.85, 0.3} 

\newcommand{\rone}{\textcolor{red}{\textbf{Dz6G}}}
\newcommand{\rtwo}{\textcolor{orange}{\textbf{wXtb}}}
\newcommand{\rthree}{\textcolor{blue}{\textbf{HLfV}}}


\begin{abstract}
We present \modelnamelong, a new geometric representation for human faces that combines B-spline volumes with learned features, and demonstrate its use as a decoder for 3D scan registration and monocular 3D face reconstruction.
Unlike existing B-spline representations with 3D control points, \modelname is parametrized by a lattice (e.g., $8 \times 8 \times 8$) of high-dimensional control features, increasing the model's expressivity.
These features define a continuous, two-stage mapping from a 3D parametric domain to 3D Euclidean space via an intermediate feature space.
First, high-dimensional control features are locally blended using the B-spline bases, yielding a high-dimensional feature vector whose first three values define a 3D base mesh.
A small MLP then processes this feature vector to predict a residual displacement from the base shape, yielding the final refined 3D coordinates.
To reconstruct 3D surfaces in dense semantic correspondence, \modelname is queried at 3D coordinates sampled from a fixed template mesh. 
Crucially, \modelname retains the local support property of traditional B-spline representations, enabling local surface editing by updating individual control features. 
We demonstrate the strengths of this representation by training transformer-based encoders to predict \modelname's control features from unstructured point clouds and monocular images, achieving state-of-the-art scan registration results compared to recent baselines.

\end{abstract}

\vspace{-1mm}
\section{Introduction}
\label{sec:introduction}

High-fidelity 3D digital humans are central to applications in virtual reality, telepresence, and immersive entertainment.
A core challenge in computer vision and graphics is developing a representation that balances computational efficiency with the expressivity required to model diverse identities, poses, and facial expressions.

Current approaches for 3D faces typically rely on explicit surfaces, most notably 3D Morphable Models (3DMMs) \cite{blanz19993dmm,li2017flame,egger2020survey}. 
These are the standard for reconstruction \cite{deng2019accurate,feng2021deca}, tracking \cite{thies2016face2face,taubner2024flowface}, speech animation \cite{danecek2023emote,zhao2024media2face}, or acting as geometry proxies for neural head avatars \cite{zielonka2023insta,qian2024gaussianavatars}.
However, 3DMMs are often limited by a fixed mesh topology and low-dimensional parameter spaces that fail to capture person-specific high-frequency details.
While learning-based non-linear models \cite{chandran2022transformer,ranjan2018coma} offer increased flexibility, they generally lack interpretability and localized control.
Conversely, implicit representations \cite{giebenhain2023nphm} provide high surface detail but lack inherent semantic correspondence and require computationally expensive isosurface extraction (e.g., Marching Cubes) for rendering and compatibility with standard graphics pipelines.

To address these limitations, we propose \modelname, a hybrid representation that combines the localized control of B-splines with the expressivity of learning-based models.
Inspired by B-spline surfaces \cite[Chapter 3]{piegl1997nurbs} with their localized control through small grids of control points, and the spline-based transformer \cite{serifi2024b}, we leverage B-spline volumes \cite{Wu2000volumegraphics}. 
However, B-spline volumes commonly reconstruct surfaces via isosurfacing.
Instead, we utilize the volume to define a continuous mapping from a fixed template mesh to the target geometry, ensuring dense semantic correspondence in the output.

Unlike traditional B-splines that rely on 3D control points, \modelname is parameterized by a lattice of learnable, high-dimensional control features.
To evaluate the model, we query the B-spline volume using vertex coordinates from a fixed template mesh.
The interpolated high-dimensional features are processed in two stages: the first three dimensions define a coarse base mesh, while the full feature vector is passed through a lightweight Multilayer Perceptron (MLP) to predict fine-scale geometric residuals.
This architecture significantly enhances expressivity compared to standard B-splines.
Crucially, it retains desirable local support properties, enabling local surface editing by updating individual control features.

We validate this representation by training transformer-based encoders to predict \modelname parameters from unstructured point clouds and monocular images.
Our experiments demonstrate state-of-the-art performance in scan registration and reconstruction compared to recent geometric and multi-view baselines.

In summary, our main contributions are:
\begin{itemize}
    \item \modelname, a novel geometric representation that parameterizes a B-spline volume with high-dimensional learnable control features.
    \item A hybrid decoding strategy that combines B-spline interpolation with a lightweight MLP to recover fine-scale geometric residuals while maintaining the interpretability and local support of traditional B-splines.
    \item A framework to predict \modelname control features from unstructured 3D head scans, demonstrating superior registration accuracy over recent learning-based methods.
\end{itemize}

\section{Related work}
\label{sec:related_work}

\paragraph{Common face representations}
3D Morphable Models (3DMMs) \cite{li2017flame, blanz19993dmm, egger2020survey} based on triangle meshes have remained the go-to method for representing human faces in both academia and industry. 3DMMs for faces represent geometry in a compressed, disentangled latent space and provide a simple linear decoder for shape reconstruction. 3DMMs continue to be used as one of the main driving representations for full head avatars \cite{zielonka2023insta,qian2024gaussianavatars,xiang2024flashavatar}. 
On the other hand, nonlinear face models \cite{ranjan2018coma, chandran2022transformer} rely on the expressivity of neural networks to represent highly detailed faces with a low dimensional latent space. The latent spaces of these neural shape models have also evolved to provide semantic controls \cite{Chandran2020, Li2020, foti20223d} similar to traditional 3DMMs. 
Another emerging paradigm for modeling facial geometries are implicit neural representations \cite{park2019deepsdflearningcontinuoussigned}. Implicit face models \cite{giebenhain2023nphm,potamias2025imhead, zheng2022imface} use MLPs to model signed distance fields (SDFs) from which a face mesh can be extracted using marching cubes. The ability to continuously evaluate these models in space makes them an attractive representation for driving full head avatars  \cite{giebenhain2024npga}.
In comparison to existing face representations, \modelname is a hybrid representation based on B-Spline volumes. \modelname allows for localized control over an underlying geometry through high-dimensional B-Spline control features, and also uses a lightweight residual MLP to boost the expressivity of the linear B-Spline bases.  CUBE can be queried at 3D coordinates sampled from a fixed template mesh, but can also be evaluated continuously like an implicit model.

\paragraph{CAD representations}
Non-uniform rational B-Splines (NURBS), on which \modelname is built, are a standard representation for designing and manipulating 3D geometries with high precision in Computer-Aided Design (CAD) \cite{Wu2000}. 
Unlike polygon-based modeling (ex. meshes), NURBS are continuous mathematical objects representing smooth surfaces that can be manipulated interactively through a sparse set of control points.
NURBS volumes \cite{Wu2000, Jianwen2001} are an extension of B-Spline surfaces for modeling continuous volumetric fields. 
For example, Li \cite{li2022nurbsvolume} proposed the use of a NURBS volume as a compact representation of a volumetric scalar field that is optimized to match multiview projection constraints. 
A shortcoming of NURBS surfaces and volumes is their inability to represent complex geometries such as face shapes with a limited number of control points. They also have limited support in standard rendering software and often have to be discretized into a polygon mesh for downstream use. 
\modelname replaces the 3D control points used in standard B-Spline volumes with high-dimensional control features, and a neural residual MLP. \modelname can therefore represent complex geometries with only a few control points, while leveraging its residual MLP to model high-frequency details. \modelname is also parameterized with an underlying template mesh, removing the need for an additional surface extraction step to produce polygon meshes and thus can be easily plugged into existing graphics pipelines. 

\paragraph{Cage-based deformation} 
Free-form geometric deformation using volumetric control lattices is a well studied technique in graphics, wherein the vertices of a mesh are obtained through an affine sum of control points in $\mathcal{R}^3$. The weights for a given vertex are determined by mapping its location to the parametric space of the control lattice. Trivariate B-Splines \cite{Griessmair89}
are one of several geometric representations that have been used for this purpose. We refer to this detailed survey \cite{Stroter24}
for a comprehensive discussion. Jung \etal \cite{Jung21}
proposed the use of such a free-form deformation representation as the output representation to regress 3D face shapes from input images. \modelname extends standard free-form deformation for geometry with two main differences. First, we propose the use of high dimensional control lattice along with a refinement MLP to increase the level of detail in the reconstructed geometry. Second, we parameterize the control lattice through a normalized template mesh that allows us to establish correspondences between the control features and known semantic locations on the template mesh.

\paragraph{Applications-specific face representations}
In this paper, we demonstrate how \modelname can be useful in practice through two applications: i) facial scan registration and ii) image-based regression. We now provide a short review of the different representations used in these applications. 

\emph{Facial scan registration:}
Scan registration deforms a standard template mesh to match a 3D face scan \cite{Bee10}. Unprocessed scans vary drastically in the number and ordering of vertices and are not amenable to statistical analysis, making registration a crucial step for practical utility. Methods are generally classified into optimization-based and learning-based approaches.
Salazar \etal \cite{salazar2014registration} proposed a method for mesh-based scan registration that automatically annotates sparse landmarks on a scan using a markov network. The annotated landmarks are then used to rigidly align a template mesh with the scan, followed by an iterative optimization with closest point constraints to guide the registration. 
Topo4D~\cite{li2024topo4d} optimizes the properties of 3D Gaussians \cite{kerbl3Dgaussians} attached to a template mesh to achieve topologically consistent face meshes and 8K textures from multi-view videos of a single subject. 

Learning-based scan registration methods deal with the challenging task of mapping an unstructured representation (i.e. scan) to a structured shape (i.e. template mesh) through a neural network. 
Prokudin \etal \cite{prokudin2019bps} introduced Basis Point Sets (BPS), encoding scans with arbitrary point counts into fixed-size embeddings using randomly distributed basis points $\in \mathbb{R} ^ 3$. The embedding consists of the computed distances from each basis point to its closest scan neighbor. This vector is then passed through an MLP to decode the registered mesh.
Liu \etal \cite{liu2019face_modeling} employ a PointNet \cite{qi2016pointnet} encoder to reduce input point clouds into disentangled identity and expression embeddings. These are transformed into their respective offsets and summed to produce the final shape.
Similarly, Shape-my-face \cite{bahri2021smf} also uses a PointNet encoder to produce identity and expression embeddings from a scan, but uses a graph-convolution-based decoder to predict the final shape. 
ImFace \cite{zheng2022imface} is a morphable model based on implicit neural representations. Trained as an auto-decoder, it predicts the signed distance field value at a 3D query point conditioned on identity and expression vectors.

\emph{Image reconstruction.} 
Most methods for monocular \cite{feng2021deca,danecek2023emote} and multi-view face reconstruction \cite{bolkart2023tempeh,li2021tofu,li2024grape,liu2022refa} use triangular meshes as their underlying representation.
VHAP~\cite{qian2024vhap} is one such approach that fits FLAME \cite{li2017flame} and an appearance prior to monocular videos or multi-view images using a differentiable renderer.
Bai \etal \cite{bai2020dfnrmvs} proposed the adaptive face model: a person specific linear shape basis that is applied on top of an underlying 3DMM and is learned from an input video through an iterative non-rigid MVS.
Wang \etal \cite{wang2025reconstructingtopologyconsistentfacemesh} use a tri-plane representation \cite{Chan2022} along with an MLP decoder to fit an implicit neural representation to multi-view images. 

\section{Method}
\label{sec:method}

\begin{figure}
    \centering
    \includegraphics[width=\linewidth]{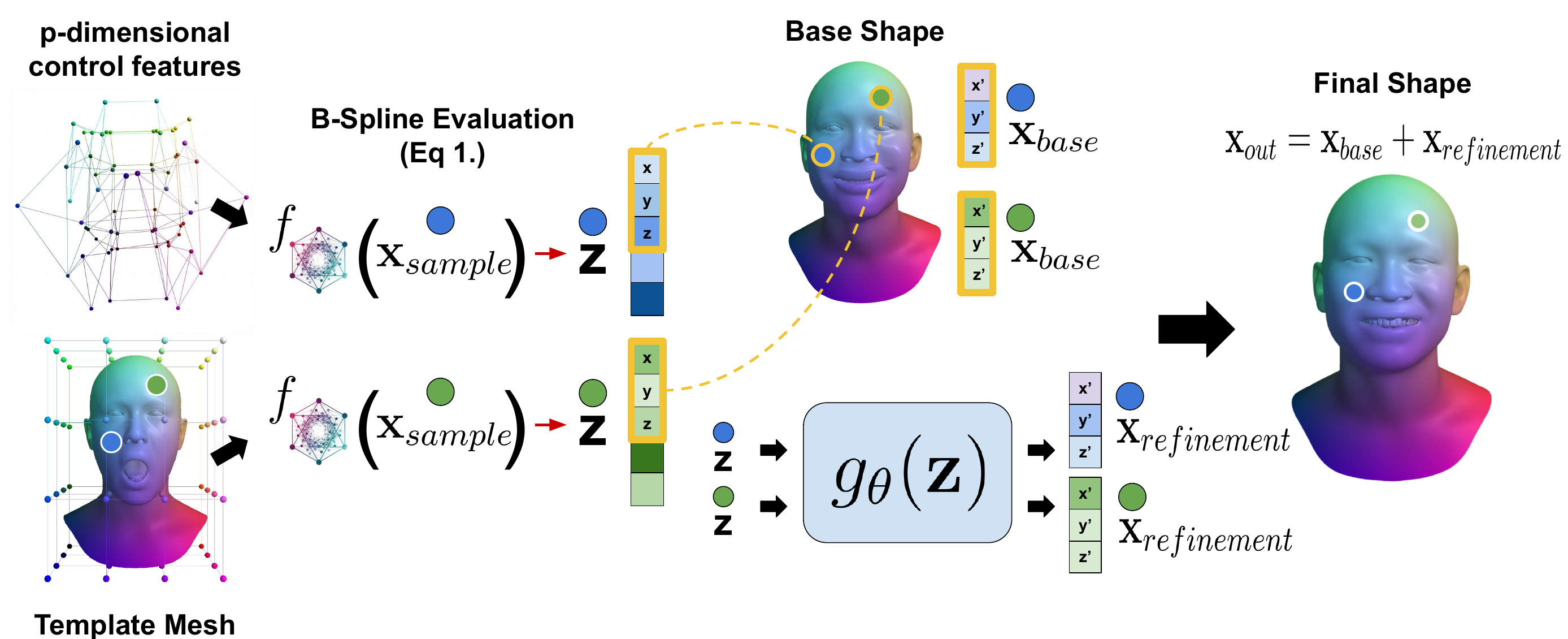}
    \caption{
    Defined by a lattice of high-dimensional control features, \modelname reconstructs a 3D face in a two-stage process.
    First, 3D coordinates are sampled from a fixed template mesh.
    These coordinates are then mapped to high-dimensional features via B-spline interpolation using the lattice of control features.
    The first three values of the resulting high-dimensional feature vector form a coarse base shape.
    Finally, the full feature vector is input to a small MLP, which predicts coordinate offsets (residuals) from the base shape, resulting in the refined 3D point coordinates.
    }
    \label{fig:concept}
\end{figure}

\begin{figure}
    \centering
    \includegraphics[width=\linewidth]{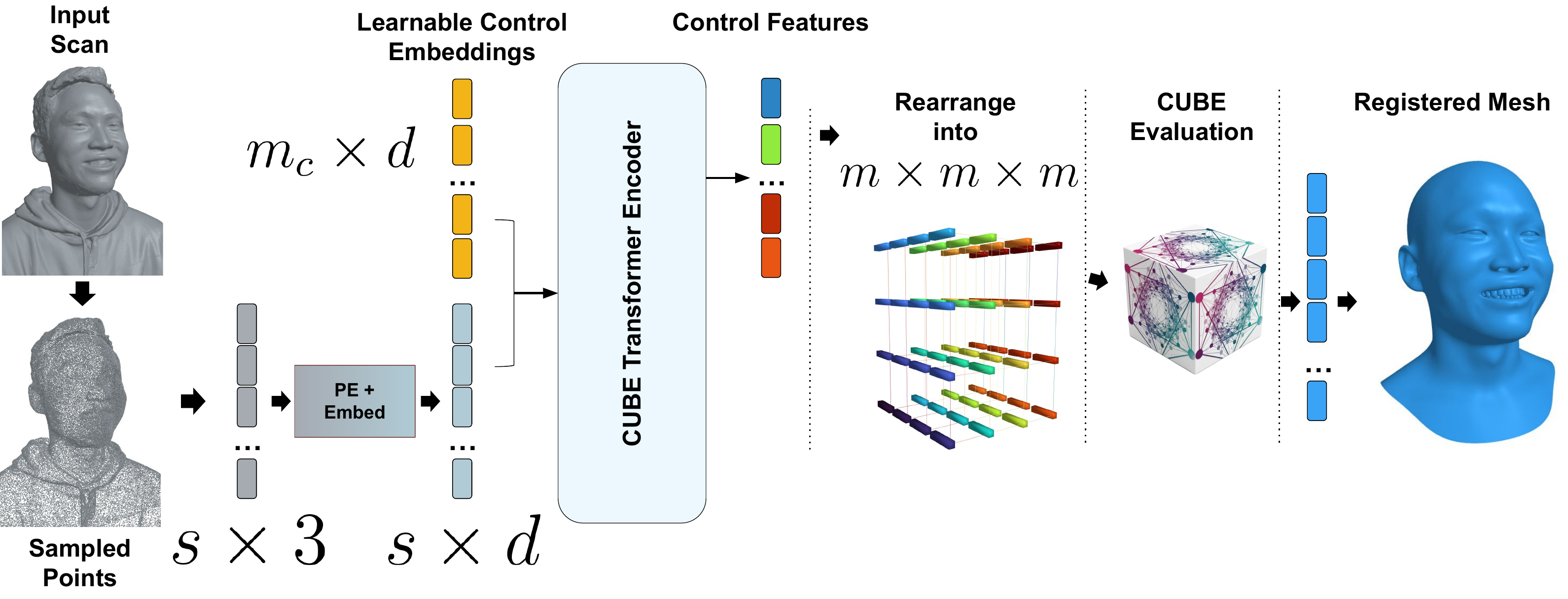}
    \caption{
    We register scans to a common mesh topology by directly predicting the control features of \modelname.
    For this, input scan vertices are tokenized, concatenated with trainable control tokens, and then passed through multiple transformer layers. 
    The resulting control token embeddings are extracted and reshaped to form the feature lattice of \modelname.
    Querying \modelname with a fixed template mesh points then reconstructs the registered 3D mesh.
    }
    \label{fig:arch}
\end{figure}

\subsection{B-spline feature volumes}
\label{subsec:nurbsoverview}

\paragraph{Representation}
A \nurbs (Non-Uniform Rational B-Spline) feature volume is a generalization of a \nurbs volume~\cite{li2022nurbsvolume,park2005volumetric,Wu2000volumegraphics}, where the $\featuredim$-dimensional \nurbs output $\nurbsfunctionparams{u,v,w}$ represents not 3D spatial coordinates, but an abstract, high-dimensional feature vector. 
Given a $(\gridsize \times \gridsize \times \gridsize)$ lattice of $\featuredim$-dimensional control features $\controlfeature{i}{j}{k} \inR{\featuredim}$, each with a control weight $\weight{i}{j}{k} \inR{}$, this \nurbs feature volume is defined as a function $\nurbsfunction: \R{3} \rightarrow \R{\featuredim}$ that maps a parameter vector $(u, v, w)$ to a $\featuredim$-dimensional feature vector.
Formally, $f$ is defined as:
\begin{equation}
    \nurbsfunctionparams{u,v,w} = \frac{
    \sum_{i,j,k=1}^{\gridsize}
    \basisfunction{i}{\degree}{u}
    \basisfunction{j}{\degree}{v}
    \basisfunction{k}{\degree}{w}
    \weight{i}{j}{k}
    \controlfeature{i}{j}{k}
    }
    {
    \sum_{i,j,k=1}^{\gridsize}
    \basisfunction{i}{\degree}{u}
    \basisfunction{j}{\degree}{v}
    \basisfunction{k}{\degree}{w}
    \weight{i}{j}{k}
    }.
    \label{eq:nurb_volume}
\end{equation}
Here, $\basis{i}{\degree}: \R{} \rightarrow \R{}$ are B-spline basis functions~\cite[Section 2.2]{piegl1997nurbs} of degree $\degree$.
The B-spline basis is defined over a sequence of non-decreasing real numbers $\knotvector = \{\knots_1,...,\knots_{\gridsize+\degree+1}\}$, called \emph{knot vector}, that define where and how control features influence the volume. 
With this knot vector, the B-spline basis function of degree $\degree$ is defined recursively as ($\degree\geq1)$:
\begin{equation*}
    \basisfunction{i}{\degree}{u} =
    \frac{u-\knots_i}{\knots_{i+\degree}-\knots_{i}} \basisfunction{i}{\degree-1}{u}
    +
    \frac{\knots_{i+\degree+1}-u}{\knots_{i+\degree+1}-\knots_{i+1}} \basisfunction{i+1}{\degree-1}{u},
\end{equation*}
with the base case ($\degree=0)$:
\begin{equation*}
    \basisfunction{i}{0}{u} := 
    \begin{cases} 
        1 & \text{if } \knots_i \le u < \knots_{i+1}, \\
        0 & \text{otherwise}.
    \end{cases}
\end{equation*}
One interesting property of the B-spline basis functions is their locality, as the $\basisfunction{i}{\degree}{u} = 0$, for parameters $u \notin \left[\knots_i, \knots_{i+\degree+1} \right)$.
This local support property propagates to the B-spline volume, where each output feature vector is influenced by at most $(\degree+1)^3$ control features.
For simplicity, the parameters $(u,v,w)$ share the same knot vector $\knotvector = \{0,...,0,\knots_{\degree+2},...,\knots_{\gridsize},1,...,1 \}$, with the first and last $\degree+1$ values fixed to zero and one, respectively, defining the parameter domain to the interval $\left[0, 1\right]$.
We choose $\degree=2$ for all experiments.

\paragraph{Background}
The \nurbs volume of Li~\cite{li2022nurbsvolume} is a special case of \Cref{eq:nurb_volume} with $\featuredim=1$, resulting in a scalar field output, from which a surface is reconstructed with isosurface extraction. 
The \nurbs feature volume generalizes the standard \nurbs surface~\cite[Chapter 4]{piegl1997nurbs} by extending its parameter domain from $(u, v)$ to $(u, v, w)$, and by representing the control features $\controlfeature{i}{j}{k}$ in an arbitrary $\featuredim$-dimensional space, rather than constraining it to $\featuredim=3$.

\subsection{\modelname}
\label{subsec:cube}

Parametrized by the $(\gridsize \times \gridsize \times \gridsize)$ lattice of $\featuredim$-dimensional control features $\controlfeature{i}{j}{k}$, \modelname reconstructs a 3D point $\outvector \inR{3}$ corresponding to a parameter point $\pointsample \inR{3}$.
As shown in \figref{fig:concept}, this is done in two stages: first, a base shape is reconstructed from the B-spline volume, before in a second stage, this is refined with an MLP.

\paragraph{Coarse base shape}
The 3D parametric point $\pointsample$ is first mapped to a $\featuredim$-dimensional feature vector $\featurevector$ through the B-spline volume $\nurbsfunctionparams{\pointsample} \rightarrow \featurevector$ (\Cref{eq:nurb_volume}).
The resulting feature vector is designed to encode both coarse shape and refinement details.
The first three dimensions of $\featurevector$ are the 3D coordinates of base point $\basepoint := \featurevector_{1:3}$.

\paragraph{Residual displacement}
The second stage refines the base point by incorporating high-frequency details encoded in the feature vector $\featurevector$.
For this, $\featurevector$ is passed through a lightweight 4-layer MLP $\mlp: \R{\featuredim} \rightarrow \R{3}$.
This MLP predicts the residual displacement $\mlpfunction{\featurevector} \rightarrow \pointrefinement$.
The final 3D point coordinates $\outvector$ is then computed as the sum of the base point and the refinement, $\outvector = \basepoint + \pointrefinement$.

\paragraph{Point sampling}
To guarantee dense semantic correspondence across different reconstructed 3D face surfaces, each defined by different control features, we query \modelname\ at fixed parametric coordinates $\pointsample$.
To reconstruct surface points, we consider the parametric coordinates $\pointsample \in [0, 1]^3$ derived from the surface of a fixed template mesh (normalized to the unit cube $\left[0, 1\right]^3$).
By querying \modelname\ at the vertex locations of the template mesh (i.e., the parameter points $\pointsample$ corresponding to those vertices), we reconstruct 3D meshes with the same mesh topology as the template mesh.
Further, by sampling \modelname\ at non-vertex locations (e.g., intermediate points on the template surface), we can reconstruct 3D points for arbitrary mesh topologies, all while maintaining dense semantic correspondence.

\subsection{Mapping scans to \modelname control features}

Scans acquired from multi-view capture setups consist of an unordered, variable number of 3D points with diverse positions and orientations in space.
Our objective is the registration of such a scan to a common mesh topology using a feed-forward prediction model.
This model leverages \modelname as the decoder by directly predicting \modelname's control features, specifically the tensor $\controlfeature{}{}{}$, as illustrated in \figref{fig:arch}.
First, we center the scan by subtracting the mean of its vertices.
Unlike prior feed-forward scan registration work \cite{prokudin2019bps}, the scans are not pre-processed into a canonical orientation.
This simplifies the model's application at test time by eliminating the need for additional pre-processing.
We optionally apply a 10-band Fourier position embedding \cite{Tancik2020} on the scan vertices.
While this does not impact the model's final accuracy, it accelerates training convergence.
The scan vertices are then tokenized into $\numscanvertices$ tokens of size $\featuredim$, where $\featuredim \gg 3$.
Additionally, we define $\totalgridsize = \gridsize \times \gridsize \times \gridsize$ trainable control tokens of size $\featuredim$.
These $\totalgridsize$ control embeddings are concatenated with the tokenized scan vertices, yielding a final sequence of input tokens of length $\totalgridsize + \numscanvertices$.
These tokens are subsequently input into a sequence of feed-forward Transformer blocks \cite{chandran2022transformer, Chandran2025} utilizing XCiT attention \cite{el2021xcit}.
Finally, at the output of the Transformer encoder, we extract the embeddings corresponding to the first $\totalgridsize$ tokens.
These extracted embeddings are then reshaped to form the $\gridsize \times \gridsize \times \gridsize$ lattice of \modelname control features.
The remaining $\numscanvertices$ scan tokens from the encoder are discarded.
The mesh in correspondence is then reconstructed by querying \modelname at the vertex locations of the normalized template mesh.
\subsection{Loss functions}
\label{subsec:lossfn}

We train our model for scan registration by jointly training the encoder and the decoder in a supervised manner.
Given a dataset (\secref{sec:implementation}) of scans and corresponding registrations, we compute an L1 loss on both the predicted base mesh $B$ and the final mesh $M$ with respect to the ground truth registration mesh.
Both losses are equally weighted. 

\section{Implementation details}
\label{sec:implementation}
\subsection{Training data}
\label{subsec:datasets}
\begin{figure}
    \centering
    \includegraphics[width=0.9\linewidth]{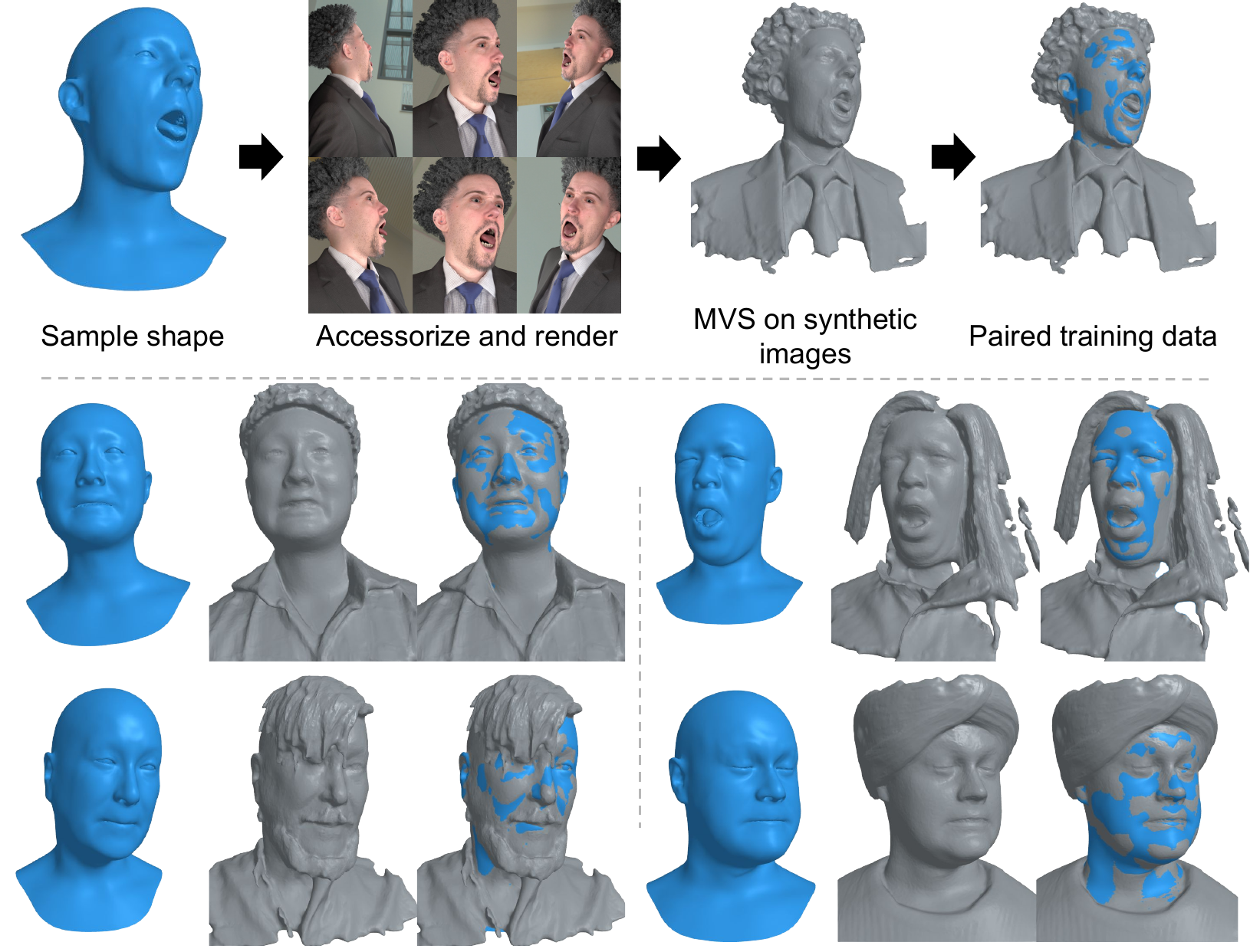}
    \caption{Starting from a randomly sampled mesh, we accessorize it with skin appearance, clothing, and hair and render it from 16 cameras using Blender. The synthetic images along with the cameras are then used to reconstruct a scan \cite{qiu2024chosen} to provide paired training data. We show some examples of mesh-scan pairs generated using this approach in the last two rows.}
    \label{fig:dataset}
\end{figure}

We train \modelname for scan registration using pairs of meshes and scans obtained from a synthetic dataset \cite{Chen2025}. Crucially, to avoid the need for manual registration, we produce the scans starting from a randomly sampled mesh using an approach shown in \cref{fig:dataset}. Our approach is identical to that of Chen \etal \cite{Chen2025} where a procedural human approach \cite{wood2021fake} was taken to first create a complete 3D human head with skin textures, clothing, hair, accessories, etc. The procedurally generated head is then rendered using Blender \cite{cycles} from multiple camera views. These synthetic images are then fed into a multi-view reconstruction pipeline \cite{qiu2024chosen} to produce their corresponding scans. Samples from our synthetic dataset are shown in \figref{fig:dataset}. We generate 300,000 examples and follow a 90:10 train:test split. 

\subsection{Model and Training Details}
\label{subsec:archdetails}
The \modelname transformer encoder in \figref{fig:arch} is based on standard Vision Transformer (ViT) \cite{dosovitskiy2020vit} sizes. We experiment with three models \modelname-S(mall), \modelname-M(edium), and \modelname-L(arge) which match the specifications of ViT-S, ViT-M, and ViT-L respectively (\#layers, \#embedding size, \#self-attention heads). 
As training augmentations, we randomly orient the input scan and sample 50,000 points from it at random to form a point cloud. We center the sampled point cloud at the origin by subtracting their mean to result in the final set of input points to our model.  
As the B-Spline bases only depend on the query point ($(u, v, w)$ and the index ($i,j,k)$ of a control point, we can evaluate their value for every sample location on the normalized template shape during model initialization. This makes the B-Spline volume evaluation extremely fast during training.  
We train our model on 16 TPUs with a batch size of 64 for 250,000 iterations. Training our biggest model \modelname-L with $16^3$ controls takes roughly 1 day.

\section{Evaluation}
\label{sec:evaluation}

\begin{figure}
    \centering
    \includegraphics[width=0.9\linewidth]{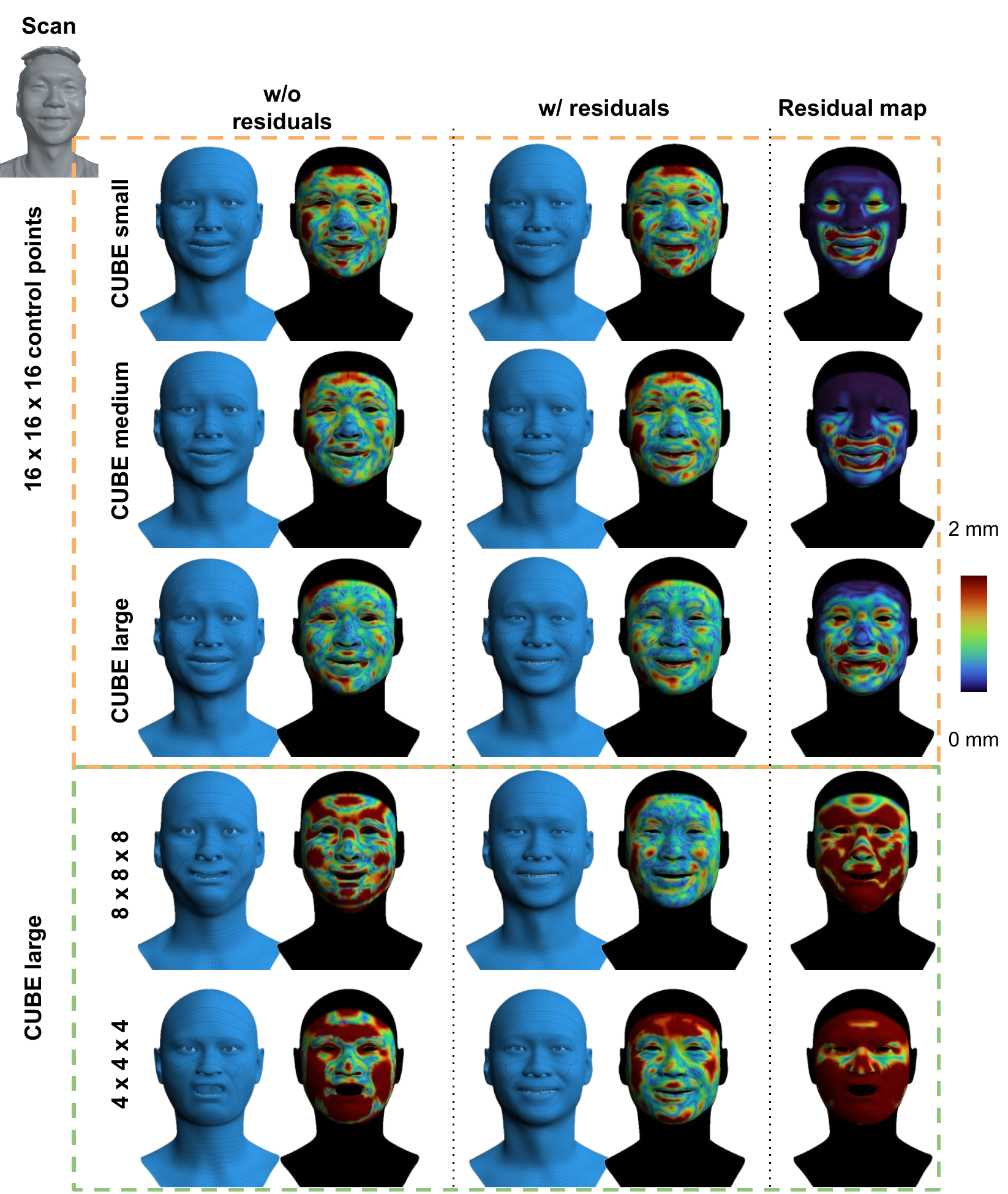}
    \caption{For a test scan shown in the top left, we evaluate \modelname models with encoders of different sizes and a varying number of control points. For each model, we show the predicted mesh and a corresponding error map visualizing the point-to-scan distance. The first columns contain shapes without applying the residuals predicted by the \modelname residual MLP, while the third and fourth columns are of shapes with the residuals applied. The last column visualizes the magnitude of the residuals across the face. The first three rows show models with 16x16x16 control points with encoders of different sizes (small, medium and large), while the last two rows are from a \modelname-Large model with $8\times8\times8$ and $4\times4\times4$ control points respectively.}
    \label{fig:modelsize}
\end{figure}

\paragraph{Ablation}
We perform an ablation study to understand the \modelname representation and analyze the impact of design choices on its representation power.  As we know from \secref{subsec:cube}, \modelname is parameterized by the number of control points ($\totalgridsize$), the dimensionality of each control point ($\featuredim$), and the MLP ($\mlp$). We use scan registration as our application of choice to analyze \modelname. We naturally also take the size of the \modelname transformer encoder into account. 

We train multiple scan registration models of different sizes on our synthetic scans dataset (see \secref{subsec:datasets}) and evaluate its performance on a test set of 30,000 synthetic scans. In \tabref{tab:residuals_ablation} we present the results of this study. We report two different metrics which are the i) point to scan distance (PTS) and the ii) vertex to vertex (V2V) distance. The point to scan distance measures the closest distance between the model's prediction and the input scan, while the vertex to vertex distance measures the difference between the predicted mesh and the ground truth registration. 
The performance of \modelname scales with the size of the transformer encoder and also the number of control points $\totalgridsize$. The shape offsets predicted by the residual MLP \mlp~improve the base shape across all model configurations. In \figref{fig:modelsize}, we can also visually observe that models with a larger encoder and a larger number of control points produce reconstructions that closely match the input scan. From the last row of \figref{fig:modelsize}, we can also see that the magnitude of the predicted residuals increases for models with a lower number of control points (such as $4^3$), highlighting the effectiveness of the \modelname residual MLP \mlp~in compensating for the limited expressivity of standard B-Spline volumes. From \tabref{tab:residuals_ablation}, we identify that a \modelname-Large encoder with ($16^3$) control points gives us the best results. Unless mentioned otherwise, all results in this paper are generated with this model. 

\begin{table}[t]
\centering
\footnotesize
\caption{\textbf{Impact of model sizes, number of control points $\totalgridsize$, and residual offsets.} Ablation study showing the performance of our different model configurations on our test set of synthetic scans. We report point-to-scan (PTS) and vertex-to-vertex (V2V) distances.}
\label{tab:residuals_ablation}
\setlength{\tabcolsep}{3.0pt}
\begin{tabular}{lc c ccc ccc}
\toprule
\multirow{2.5}{*}{\textbf{\shortstack{Encoder\\Size}}} & \multirow{2.5}{*}{\textbf{$\totalgridsize$}} & \multirow{2.5}{*}{\textbf{MLP}} & \multicolumn{3}{c}{\textbf{PTS (mm)}} & \multicolumn{3}{c}{\textbf{V2V (mm)}} \\
\cmidrule(lr){4-6} \cmidrule(lr){7-9}
 & & & Mean & Median & Std & Mean & Median & Std \\
\midrule
\multirow{6}{*}{\shortstack{\modelname-S\\($\featuredim=384$)}} 
    & \multirow{2}{*}{$4^3$} 
        & w/o & 2.59 & 1.70 & 2.77 & 3.66 & 3.04 & 2.34 \\
        & & w/  & 1.77 & 0.99 & 2.32 & 2.19 & 1.92 & 1.24 \\
    \cmidrule(lr){2-9}
    & \multirow{2}{*}{$8^3$} 
        & w/o & 2.16 & 1.35 & 2.49 & 2.70 & 2.30 & 1.66 \\
        & & w/  & 1.73 & 0.94 & 2.33 & 2.04 & 1.78 & 1.18 \\
    \cmidrule(lr){2-9}
    & \multirow{2}{*}{$16^3$} 
        & w/o & 1.74 & 0.95 & 2.31 & 2.09 & 1.81 & 1.24 \\
        & & w/  & \textbf{1.64} & \textbf{0.87} & \textbf{2.27} & \textbf{2.01} & \textbf{1.76} & \textbf{1.17} \\
\midrule
\multirow{6}{*}{\shortstack{\modelname-M\\($\featuredim=512$)}} 
    & \multirow{2}{*}{$4^3$} 
        & w/o & 2.62 & 1.72 & 2.81 & 3.59 & 2.94 & 2.36 \\
        & & w/  & 1.68 & 0.89 & 2.31 & 1.95 & 1.71 & 1.12 \\
    \cmidrule(lr){2-9}
    & \multirow{2}{*}{$8^3$} 
        & w/o & 2.16 & 1.35 & 2.49 & 2.61 & 2.22 & 1.61 \\
        & & w/  & 1.68 & 0.88 & 2.31 & \textbf{1.83} & \textbf{1.59} & \textbf{1.06} \\
    \cmidrule(lr){2-9}
    & \multirow{2}{*}{$16^3$} 
        & w/o & 1.70 & 0.89 & 2.32 & 1.93 & 1.66 & 1.18 \\
        & & w/  & \textbf{1.63} & \textbf{0.84} & \textbf{2.29} & 1.87 & 1.62 & 1.12 \\
\midrule
\multirow{6}{*}{\shortstack{\modelname-L \\ ($\featuredim=1024$)}}
    & \multirow{2}{*}{$4^3$} 
        & w/o & 2.63 & 1.73 & 2.80 & 3.62 & 3.01 & 2.30 \\
        & & w/  & 1.62 & \cellcolor{tabsecond}{0.81} & 2.32 & 1.81 & 1.58 & 1.06 \\
    \cmidrule(lr){2-9}
    & \multirow{2}{*}{$8^3$} 
        & w/o & 2.04 & 1.21 & 2.47 & 2.44 & 2.06 & 1.55 \\
        & & w/  & \cellcolor{tabsecond}{1.47} & \cellcolor{tabfirst}{0.67} & \cellcolor{tabsecond}{2.27} & \cellcolor{tabfirst}{1.50} & \cellcolor{tabfirst}{1.27} & \cellcolor{tabfirst}{0.94} \\
    \cmidrule(lr){2-9}
    & \multirow{2}{*}{$16^3$} 
        & w/o & 1.65 & \cellcolor{tabsecond}{0.81} & 2.33 & 1.78 & 1.52 & 1.11 \\
        & & w/  & \cellcolor{tabfirst}{1.40} & \cellcolor{tabfirst}{0.67} & \cellcolor{tabfirst}{1.63} & \cellcolor{tabsecond}{1.54} & \cellcolor{tabsecond}{1.31} & \cellcolor{tabsecond}{0.96} \\
\bottomrule
\end{tabular}
\end{table}

\paragraph{Quantitative evaluation}
Next, we quantitatively compare \modelname against two feed-forward scan registration baselines.
Our first baseline is the Basis Point Set (BPS) method \cite{prokudin2019bps}, which requires only the raw scan. BPS encodes scans with a variable number of points into fixed-size embeddings using distances to a set of random basis points; these are then processed by an MLP to predict the mesh. For our evaluation, we vary the model capacity by training three BPS models with 512, 1024, and 2048 basis points respectively. Following \cite{prokudin2019bps}, we sample these points from a uniform sphere tightly enclosing the data.
We also compare against TEMPEH \cite{bolkart2023tempeh}, a state-of-the-art registration method utilizing multiview images. In contrast to BPS and \modelname, TEMPEH does not take the 3D scan as input. Instead, it extracts per-pixel image features and fuses them across views into a 3D feature volume. The feature volume is processed by a 3D convolutional network that predicts a 'Global' estimate of the face shape. This is followed by a refinement step, where a local volume of features in the neighborhood of every vertex is processed independently by a refinement network to further localize its position in space.

For a fair comparison, we train all three methods BPS, TEMPEH and \modelname, on the same training dataset (\secref{subsec:datasets}). We test these models on 18,000 scans derived from captures of real humans in a multi-view setup \cite{qiu2024chosen}. Our real test set contains a wide range of identities performing a variety of facial expressions. We also run a standard offline registration pipeline to obtain ground truth registrations for these captures.

\begin{table}[t]
\centering
\footnotesize
\caption{\textbf{Comparison with State-of-the-Art Methods.} Quantitative results for BPS \cite{prokudin2019bps}, TEMPEH \cite{bolkart2023tempeh}, and \modelname across different model sizes and residual configurations. We report the point-to-scan (PTS) and vertex-to-vertex (V2V) distances.}
\label{tab:sota_comparison}
\setlength{\tabcolsep}{3.5pt}
\begin{tabular}{l ccc ccc}
\toprule
\multirow{2.5}{*}{\textbf{Method}} & \multicolumn{3}{c}{\textbf{PTS (mm)}} & \multicolumn{3}{c}{\textbf{V2V (mm)}} \\
\cmidrule(lr){2-4} \cmidrule(lr){5-7}
 & Mean & Median & Std & Mean & Median & Std \\
\midrule
\textbf{Ground Truth} & 0.60 & 0.40 & 0.72 & n/a & n/a  & n/a \\
\midrule
\textbf{BPS \cite{prokudin2019bps}} & & & & & & \\
\hspace{3mm} 512 points  & 4.90 & 4.33 & 3.35 & 14.84 & 14.72 & 3.92 \\
\hspace{3mm} 1024 points & 4.66 & 4.01 & 3.30 & 15.16 & 15.03 & 3.81 \\
\hspace{3mm} 2048 points & 5.03 & 4.21 & 3.69 & 17.78 & 17.79 & 3.87 \\
\midrule
\textbf{TEMPEH \cite{bolkart2023tempeh}} & & & & & & \\
\hspace{3mm} w/o Refinement & 1.45 & 1.15 & 1.10 & 2.46 & 2.24 & 1.26 \\
\hspace{3mm} w/ Refinement  & \cellcolor{tabsecond}{1.13} & \cellcolor{tabsecond}{0.80} & 0.98 & 2.27 & 2.01 & 1.27 \\
\midrule
\textbf{(\modelname-S)} & & & & & & \\
\hspace{3mm} w/o residuals & 1.24 & 0.96 & 0.99 & 2.20 & 1.91 & 1.30 \\
\hspace{3mm} w/ residuals  & 1.17 & 0.90 & 0.93 & 2.13 & 1.86 & 1.24 \\
\midrule
\textbf{(\modelname-M)} & & & & & & \\
\hspace{3mm} w/o residuals & 1.18 & 0.89 & 0.97 & 2.05 & 1.76 & 1.24 \\
\hspace{3mm} w/ residuals  & \cellcolor{tabsecond}{1.11} & 0.85 & \cellcolor{tabsecond}{0.91} & 1.99 & 1.72 & 1.19 \\
\midrule
\textbf{(\modelname-L)} & & & & & & \\
\hspace{3mm} w/o residuals & 1.12 & 0.82 & 0.96 & \cellcolor{tabsecond}{1.90} & \cellcolor{tabsecond}{1.63} & \cellcolor{tabsecond}{1.17} \\
\hspace{3mm} w/ residuals  & \cellcolor{tabfirst}{0.95} & \cellcolor{tabfirst}{0.71} & \cellcolor{tabfirst}{0.82} & \cellcolor{tabfirst}{1.69} & \cellcolor{tabfirst}{1.45} & \cellcolor{tabfirst}{1.02} \\
\bottomrule
\end{tabular}
\end{table}

\begin{figure}
    \centering
    \includegraphics[width=\linewidth]{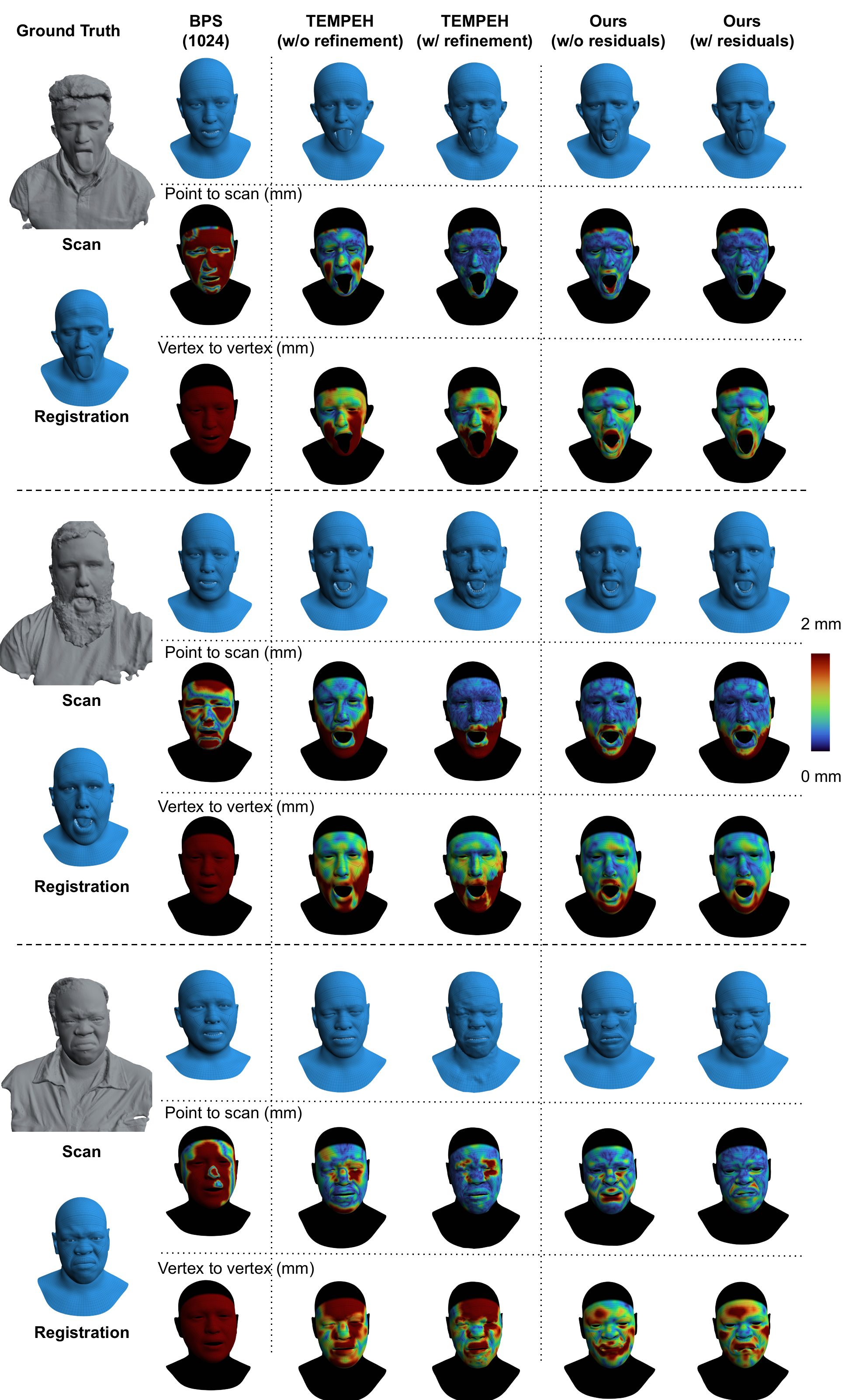}
    \caption{We show qualitative comparisons of BPS \cite{prokudin2019bps}, TEMPEH \cite{bolkart2023tempeh} and \modelname for predicting registered meshes from three different real scans in various expressions. BPS struggles with scans of varying orientations and collapses to predicting an average face shape that only rigidly matches the location of scan vertices. While TEMPEH-Refined improves on the predictions of its Global stage, it produces noisy reconstructions with geometric artifacts (last example). In contrast, \modelname produces reconstructions with the lowest errors, while matching the identity and expression of the scan closely even for challenging expressions (ex: a person having their tongue out in the first row). }
    \label{fig:comparehero}
\end{figure}

As we see in \tabref{tab:sota_comparison}, our \modelname model for scan registration outperforms previous baselines for feed-forward facial scan registration. We observe that BPS is sensitive to the orientation of the scan and can collapse to average predictions for some scan orientations. TEMPEH-Refined, while achieving metrics comparable to \modelname-S, produces noisy reconstruction. \modelname-M and \modelname-L significantly outperform BPS and TEMPEH in our evaluations. 
\section{Applications}
\label{sec:discussion}

\paragraph{Localized editing}
A key advantage of \modelname is that its control features are localized within the B-Spline volume. \modelname readily allows a user to perform local edits on a shape by directly modifying its control features. In \figref{fig:teaser} (left), we show how \modelname's control features can be swapped across two different expressions represented with \modelname. We replace the control features in the lower half of the volume with those derived from a different expression.
\modelname also allows for the editing of individual control points to make precise edits to shapes. In \figref{fig:teaser} (middle), we show how displacing a single control feature shown in red, influences the predicted shape locally. To make this local edit, we modify only the first three dimensions of a control feature that corresponds to its location in Euclidean space \figref{fig:concept}. The rest of the control feature is left unchanged. As we see in \figref{fig:teaser} (middle), modifying one such control point in this manner only influences a local region (in this case, the lower lip). We highlight that \modelname's control features are local by design and do not require any locality-inducing losses or particular types of training data to enable localized control. 

\begin{figure}
    \centering
    \includegraphics[width=\linewidth]{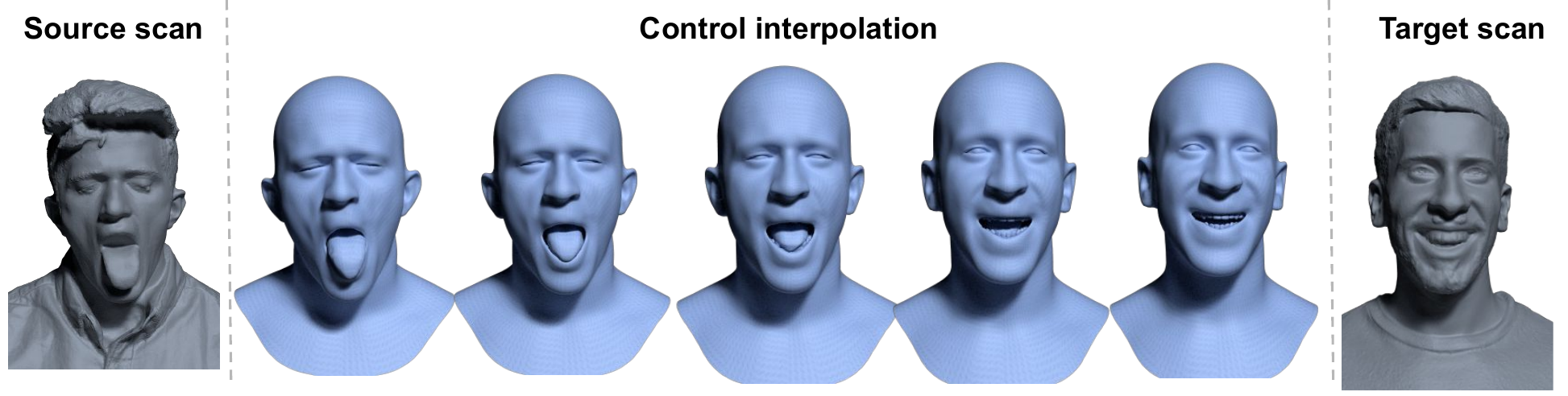}
    \caption{Control features derived from a source and a target scan are linearly interpolated and then evaluated through our \modelname decoder. The shapes resulting from the interpolated control features transition smoothly from the source to the target.}
    \label{fig:interpolation}
\end{figure}

\paragraph{Control interpolation}
As \modelname's control features have geometric meaning, they interpolate smoothly. In \figref{fig:interpolation}, we encode a source and target scan independently through the \modelname transformer encoder and obtain two sets of control features. We linearly interpolate the control features and evaluate \modelname following \figref{fig:concept} at each interpolated time step to produce the interpolated shapes. In \figref{fig:interpolation}, we see that as the interpolation progresses, the interpolated shape gradually changes in identity and expression from the source to the target, even for extreme facial expressions.  

\begin{figure}
    \centering
    \includegraphics[width=\linewidth]{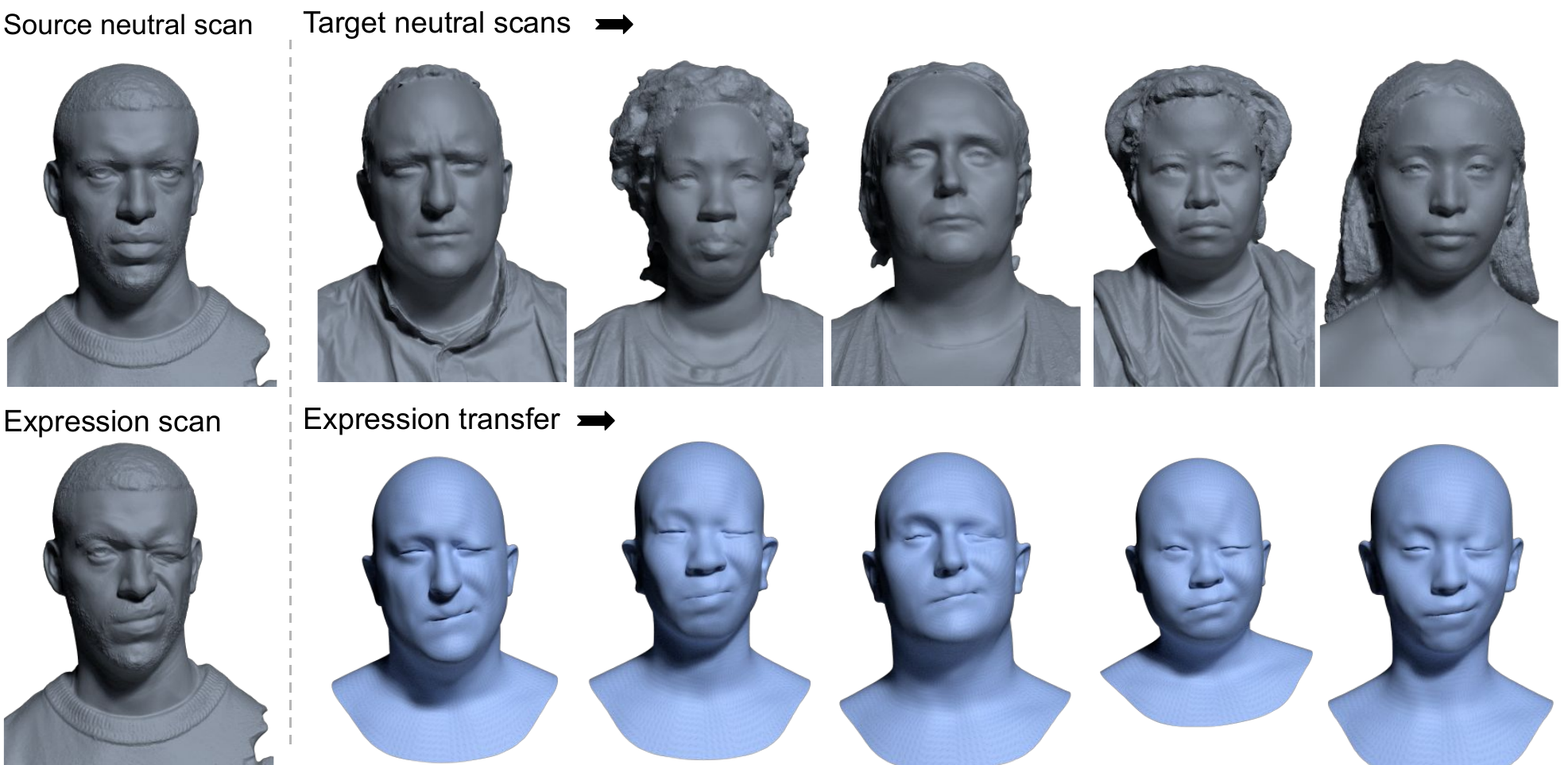}
    \caption{We compute the difference between the control features of a neutral and expression scan of a source subject (first column). This feature difference is then added to control features derived from neutral scans obtained of different target subjects (columns 2-6). In this case, \modelname control arithmetic transfers facial expressions from the source to the target subjects.}
    \label{fig:exptransfer}
\end{figure}

\paragraph{Expression transfer}
We can also perform arithmetic operations on our control features to enable applications such as expression transfer from raw scans. Given two scans of the same person (source) in two different expressions, we obtain control features corresponding to both scans from our encoder. We then compute the difference between the control features, apply it as an offset to control features obtained from a neutral scan of a target subject. In this case, control feature arithmetic results in transferring expressions from the source to target subjects as we can see in \figref{fig:exptransfer}. 

\begin{figure}
    \centering
    \includegraphics[width=\linewidth]{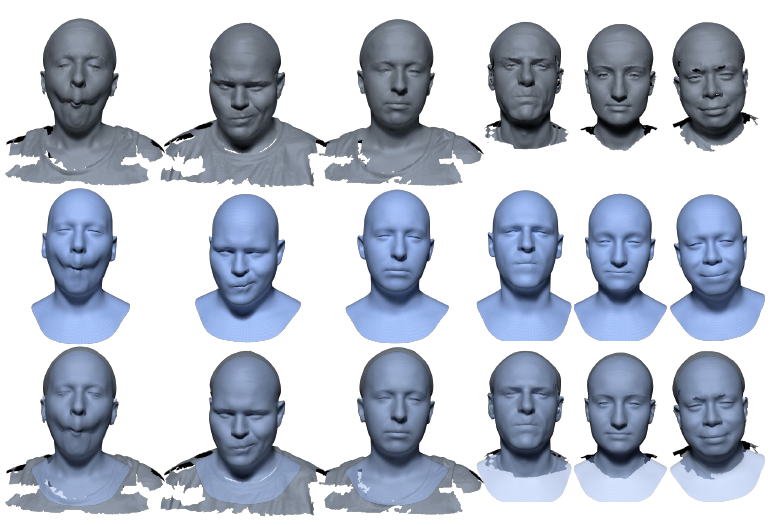}
    \caption{The \modelname scan encoder generalizes to scans captured by different setups. Here we show scans from two different datasets CoMA \cite{ranjan2018coma} (columns 1-3) and FaMoS \cite{bolkart2023tempeh} (columns 4-6). The input scans are shown in the first row, the predictions from our \modelname-L model are shown in the second row, and an overlay of the predictions with the input scan is shown in the third row.}
    \label{fig:coma_famos}
\end{figure}

\paragraph{Generalization to in-the-wild scans}
Our method generalizes well to in-the-wild scans produced by other capture setups and MVS algorithms without finetuning. In \cref{fig:coma_famos}, we run our \modelname-L model on scans produced by two public datasets; CoMA \cite{ranjan2018coma} and FaMoS \cite{bolkart2023tempeh}. Pre-processing is limited to centering each scan at the origin.

\begin{figure}
    \centering
    \includegraphics[width=0.9\linewidth]{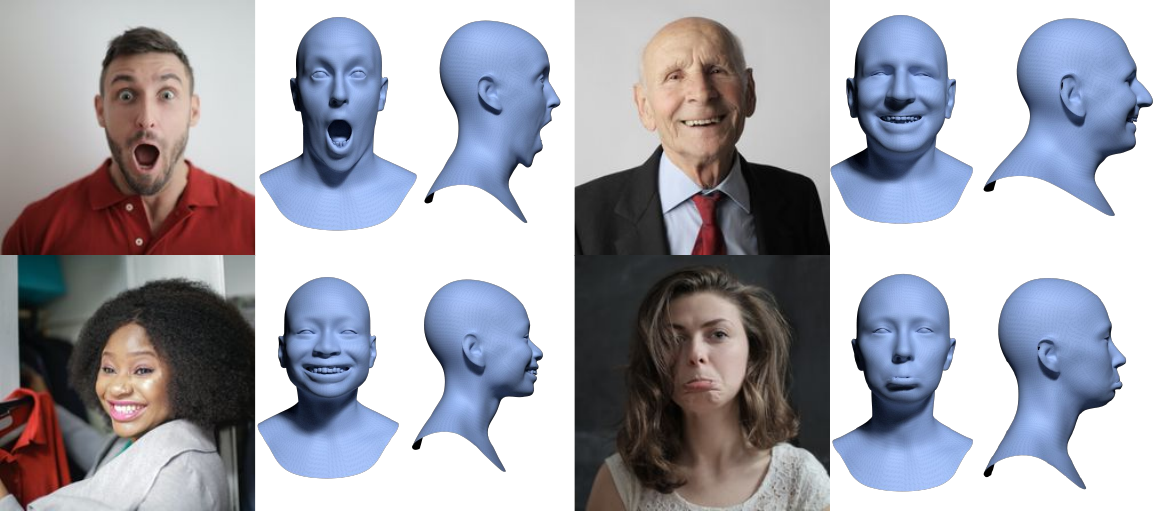}
    \caption{\modelname can also be leveraged for image understanding tasks like in-the-wild face reconstruction. \modelname is versatile and expressive enough to capture a wide range of identities and expressions without relying on an underlying 3DMM.}
    \label{fig:faceproxy}
\end{figure}

\paragraph{Image-based regression}
\label{subsec:imagereg}
By introducing an additional patchify layer before our \modelname transformer encoder in \figref{fig:arch}, we can process tokenized image patches instead of a point cloud, resulting in a ViT-like encoder with a \modelname latent space. With this simple adaptation, we can leverage the flexibility of \modelname representations for image-based regression. In \figref{fig:faceproxy}, we show results of training an image-based \modelname transformer encoder for a face reconstruction from in-the-wild images. More details and results can be found in our supplemental material. 

\section{Conclusion}
\label{sec:conclusion}

We presented \modelname, a novel geometric representation for 3D faces, which successfully hybridizes traditional B-spline volumes with high-dimensional learned control features.
\modelname is parameterized by a lattice of high-dimensional control features, and to reconstruct a 3D surface, it is continuously evaluated on the surface of a fixed template mesh.
This continuous evaluation ensures dense semantic correspondence in the resulting 3D face meshes and enables the reconstruction of different mesh topologies simply by adjusting the point sampling.
We present a transformer-based model to regress the \modelname representation from unstructured point clouds and portrait images.
When predicted by our transformer-based encoder, the \modelname representation achieves state-of-the-art accuracy in feed-forward facial scan registration.
By leveraging B-spline basis functions, \modelname's control features have localized control over the surface, allowing for the local editing of the reconstructed meshes by manipulating individual control features.
Overall, we demonstrated that \modelname is a versatile representation, with promising properties that can benefit future applications in the modeling and animation of digital humans.

{
    \small
    \bibliographystyle{ieeenat_fullname}
    \bibliography{bibliography}
}

\clearpage
\setcounter{section}{0}
\setcounter{figure}{0}
\setcounter{table}{0}
\setcounter{equation}{0}

\renewcommand{\thesection}{S\arabic{section}}
\renewcommand{\thefigure}{S\arabic{figure}}
\renewcommand{\thetable}{S\arabic{table}}
\renewcommand{\theequation}{S\arabic{equation}}

\twocolumn[{%
\begin{center}
    {\Large \bf Supplemental: Representing 3D Faces with Learnable B-Spline Volumes \par}
    \vspace{1.5em}
    {\large Prashanth Chandran \quad Daoye Wang \quad Timo Bolkart \par}
    \vspace{0.5em}
    {Google \par}
    \vspace{0.5em}
    {\tt\small \{prchandran, daoye, tbolkart\}@google.com \par}
    \vspace{2.5em}
\end{center}
}]
\section{Additional Implementation Details}
\label{sec:impdetails}

\subsection{Image-based Regression}
\label{sec:imgreg}
\begin{figure}
    \centering
    \includegraphics[width=\linewidth]{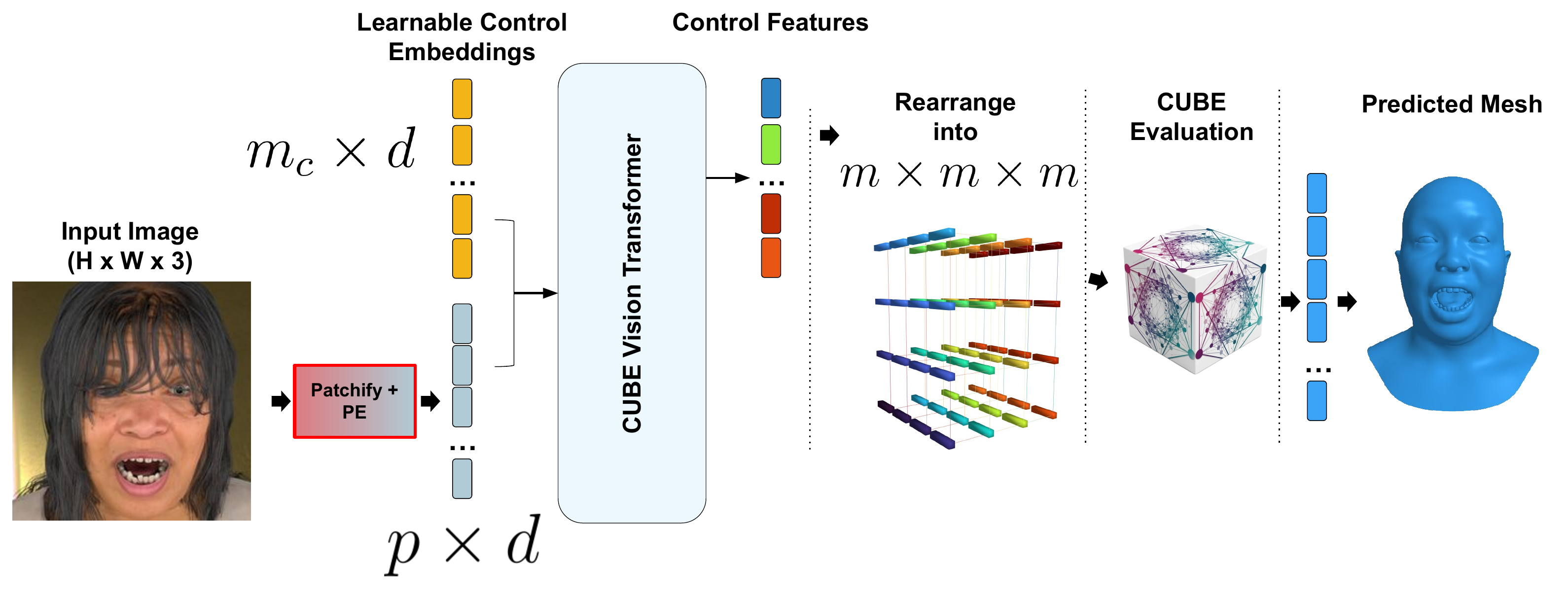}
    \caption{\modelname control features can be regressed from images by concatenating  $m_c$ learnable control embeddings to image patch tokens, and processing them with a standard vision transformer. With the exception of the \emph{patchify} layer before the transformer encoder, our architecture for image-based regression is identical to the model we introduced for scan registration in the main paper. For image-based regression, we used a ViT-Large backbone with $8 \times 8 \times 8$ control tokens.}
    \label{fig:imgarch}
\end{figure}

\paragraph{Architecture} As we briefly mentioned in the main paper, \modelname can also be used as the output representation when regressing 3D faces from an RGB image. To achieve this, we pass an input image $(H \times W \times 3)$ through a standard \emph{patchify} layer and obtain patch tokens of shape $(p \times d)$. A learnable position encoding is applied to these patch tokens \cite{dosovitskiy2020vit}.  Similar to the \modelname scan encoder, we append $m_c$ learnable control token embeddings; each of size $d$, along the sequence length of the $p$ position encoded patch tokens. We then use a standard vision transformer \cite{dosovitskiy2020vit} to regress \modelname control features from the input collection of $(p + m_c)$ tokens as shown in \cref{fig:imgarch}. The vision transformer outputs the control features that are evaluated using \modelname as shown in figure 2 of the main paper. 

\paragraph{Dataset} We use a synthetic dataset of 500K portraits rendered at a resolution of $224 \times 224$ to train our image-to-\modelname model. This synthetic dataset was generated using the same process described in section 4.1 of the main paper. As we are regressing the shape directly from a single input image, we only render an accessorized 3D head from a single randomly placed camera and skip the scan reconstruction step.  

\paragraph{Training} We use a ViT-Large model as our encoder and initialize its weights from the official MAE's encoder checkpoint \cite{MaskedAutoencoders2021} to speed up convergence. We use $8 \times 8 \times 8$ control features for \modelname. We apply standard photometric and geometric augmentations on the input images during training. We train our model with an L1 loss on the predicted shape. We use a batch size of 128 and train the model for 100,000 steps on 8 TPUs using the AdamW optimizer \cite{adamw} and a learning rate of 1e-4. 

\section{Additional Results}
\label{sec:addresults}

\subsection{Performance-cost trade off}
Table~1 of the main paper confirms a tradeoff between control point density and the level of detail added by the residual mlp $\mlp$.
With fewer control points ($4^3$), the difference in the point-to-scan distance between predictions w/ and w/o the refinement MLP is 3-7x larger, depending on the encoder size.
The inference times (measured on a V100 GPU) of our models with a varying number of control points ranges from 0.20s (CUBE-S, $4^3$) to 2.92s (CUBE-L, $16^3$). 

\subsection{Sampling robustness for scans}
\begin{figure}
    \centering
    \includegraphics[width=\linewidth]{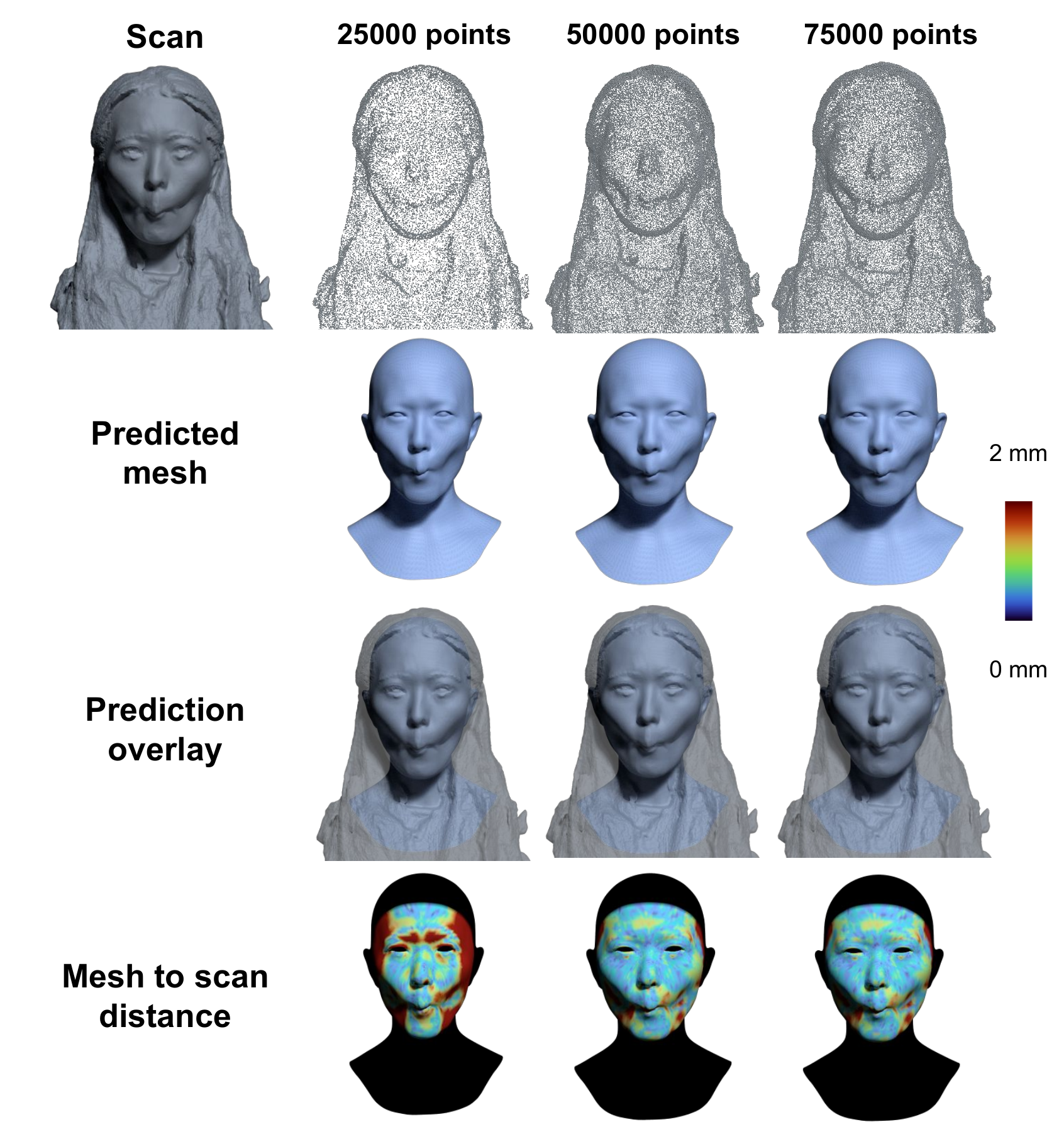}
    \caption{Given a high resolution scan, we randomly sample 25K, 50K and 75K points from the scan to result in point clouds of different sizes as seen in the first row. These sampled point clouds are independently processed by our \modelname-L model to result in the predictions seen in the second row. The third row overlays the predicted meshes on the original scan. The last row shows the mesh to scan distance for the predicted meshes. The predicted shapes remain plausible and capture the geometry of the scan in all cases demonstrating that our model is able to gracefully handle point clouds of different sizes.}
    \label{fig:sampling}
\end{figure}

Our model for scan registration is robust to varying sample counts and sampling strategies.  In \cref{fig:sampling}, we sample the same input scan three times to obtain point clouds of different sizes consisting of 25,000, 50,000 and 75,000 points respectively. The sampled point clouds are processed with \modelname-L model with $16 \times 16 \times 16$ control points  to obtain the corresponding registered meshes. As we see in \cref{fig:sampling}, the \modelname scan registration model produces perceptually similar outputs with only minor differences in face shape. 

\subsection{Effect of position encoding}
\begin{figure}
    \centering
    \includegraphics[width=\linewidth]{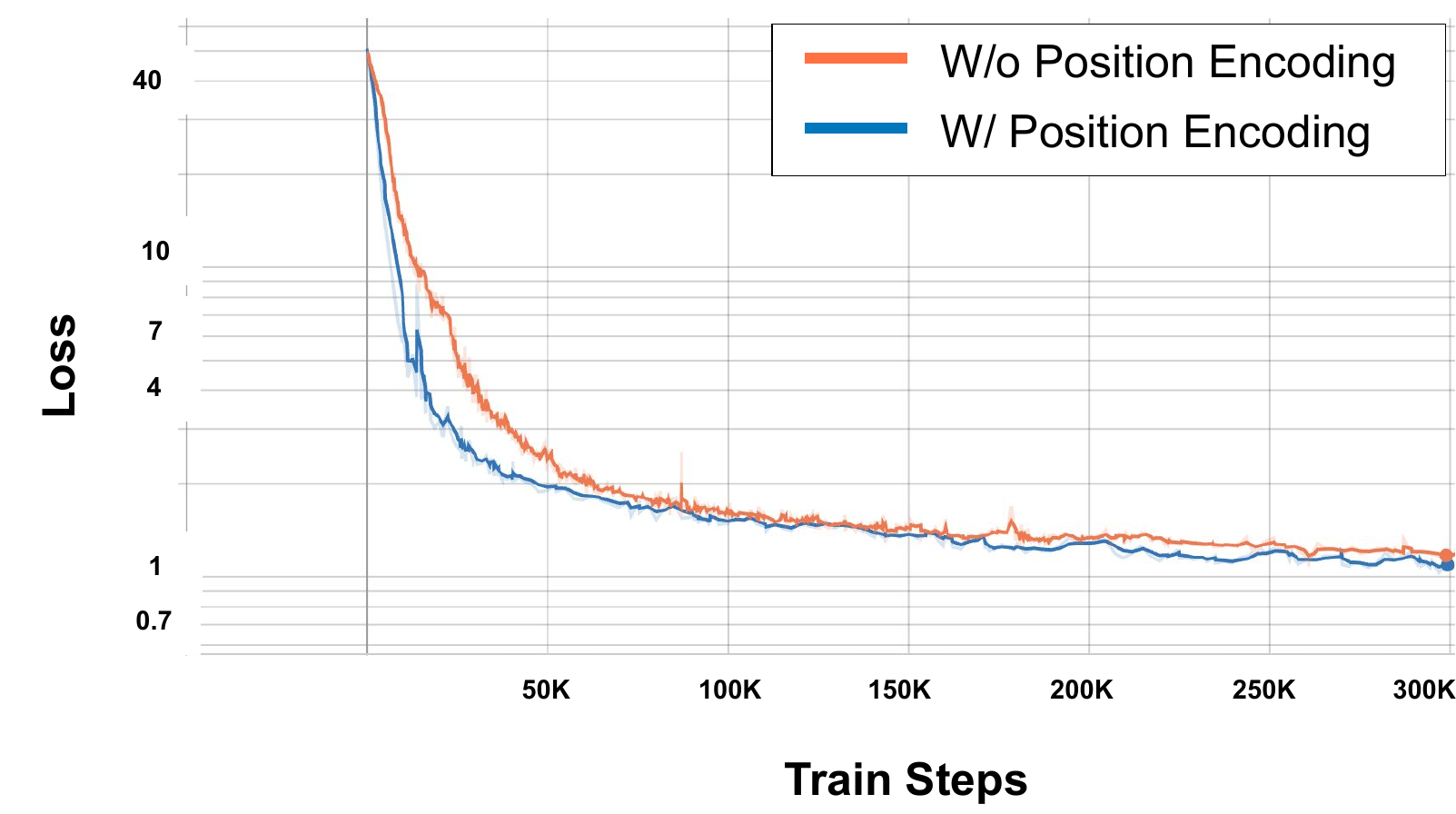}
    \caption{Position encoding (PE) the input scan points has a positive effect on the convergence of our \modelname encoder. While the difference in final training error is not significant, PE did seem to accelerate convergence in the earlier iterations of training.}
    \label{fig:pe}
\end{figure}

In \cref{fig:pe}, we compare the training curves of two identical scan-to-mesh models w/ and w/o position encoding (PE) \cite{Tancik2020} to show its effect on the convergence. We used a \modelname-Large model with $16 \times 16 \times 16$ controls for this experiment. While we did not observe a significant difference in the final training error, we noticed that PE speeds up the convergence of our model in the earlier stages of training. 

\subsection{Processing scans with colors}
Scans captured from multiview setups often contain additional information such as the color for each vertex in the mesh. These per-vertex colors may provide semantic textural cues for scan registration, allowing a model to disambiguate regions in the scan devoid of geometric features. To validate this hypothesis, we train a \modelname scan encoder to take color as an additional input for each sampled scan point. The sampled point positions and their corresponding colors are positionally encoded separately and concatenated before being fed as input to the \modelname transformer encoder. We find that providing the scan colors as an additional input lowers the reconstruction error by  roughly 12\% on our test set of 18,000 real scans.

\subsection{Topology changes}
\begin{figure}[!t]
    \centering
    \includegraphics[width=0.8\linewidth]{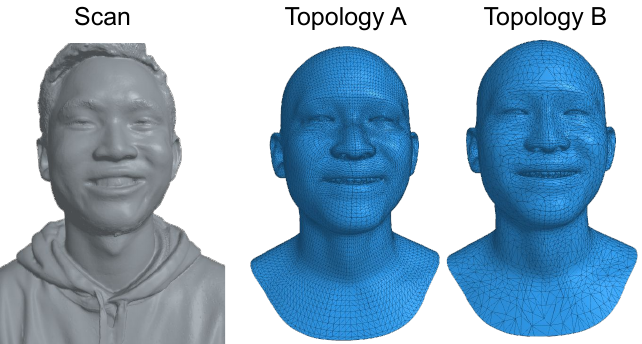}
    \caption{We show an example where by sampling \modelname at different locations on the template mesh surface, we can dictate the topology of the output shape. When combined with our transformer based encoder for scan registration, this allows \modelname to be used not only with input scans with a varying number of points, but also to represent output meshes in different topologies.}
    \label{fig:topo}
\end{figure}

\modelname is a hybrid representation of geometry that can be evaluated continuously like an implicit representation, while still maintaining correspondence with a template mesh. Therefore \modelname can produce meshes in arbitrary output topologies without any retraining. As we saw in figure 2 of the main paper, \modelname is evaluated in two stages. First, we evaluate a B-Spline volume using the predicted high dimensional control features at locations sampled from a normalized template mesh. This is followed by evaluating a point-wise residual MLP to obtain residual geometric details. By changing the topology of the template shape at inference, we can control the topology in which the output mesh is generated without having to re-train \modelname. In \cref{fig:topo}, we show an example of a predicting a registered mesh in two different output topologies from an input scan using the same model. This flexibility makes \modelname a very useful geometric representation in practice. 

\subsection{Optimizing \modelname representations}
\modelname as a representation can also be optimized or fit to target constraints without an encoder. We present two examples for optimization-based fitting with \modelname. 

\begin{figure}[t]
    \centering
    \includegraphics[width=\linewidth]{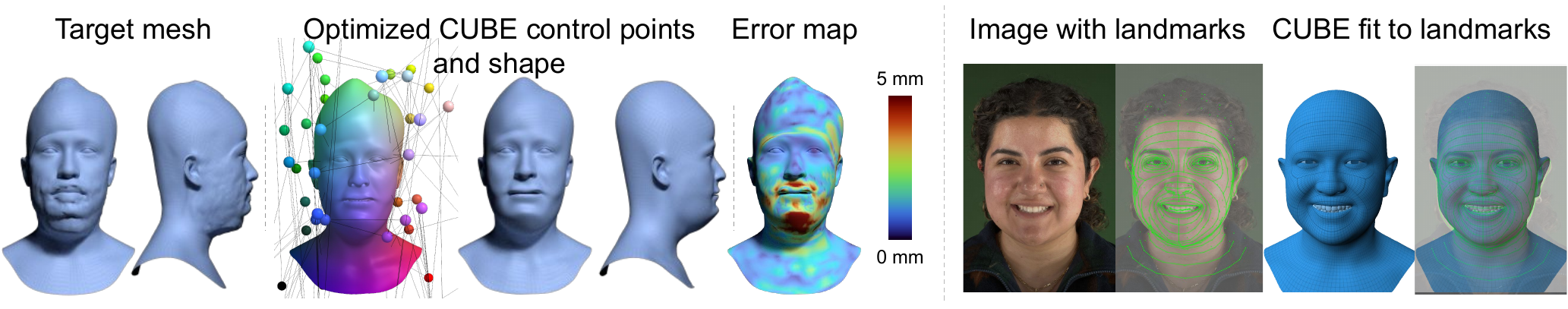}
    \caption{Left: We show how \modelname can be optimized to match the shape of a target mesh with proxy surfaces for the scalp and facial hair. The optimized control points along with the final mesh and the approximation error are also shown. Right: We show the result of fitting \modelname to match detected dense landmarks on an input image. \modelname can reasonably approximate the facial geometry with a small number of $4 \times 4 \times 4$ control features in both cases.}
    \label{fig:fitting}
\end{figure}

\paragraph{Fitting to a mesh} Given a target mesh with proxy surfaces for the scalp and the beard (see \cref{fig:fitting} left), our objective is to optimize for \modelname's latent control features $\controlfeature{i}{j}{k}$ along with the residual MLP $\mlp$ to approximate this target mesh as closely as possible. We use $4 \times 4 \times 4$ control features for this experiment where the dimension of each control feature is 16. We randomly initialize the \modelname control features and the MLP $\mlp$ and optimize them to minimize the vertex to vertex distance to the target scan using gradient descent. We use the adamw \cite{adamw} optimizer with a learning rate of 1e-3 and optimize for 2000 steps. As seen in \cref{fig:fitting} (left), \modelname can approximate the facial geometry with proxy surfaces with only a small number of control points. 

\paragraph{Fitting to 2D landmarks} 
We can optimize \modelname to fit detected 2D landmarks on images using a standard landmark re-projection energy (see \cref{fig:fitting} right). Given a portrait image of a subject, we run a dense landmark detector to detect ~600 landmarks in the image. We optimize for the \modelname control features in camera space so that the recovered shape when projected on the image, matches the detected 2D landmarks. We assume that the intrinsics are fixed for this experiment. The optimization parameters are the same as for the mesh fitting experiment, and we solve for $4 \times 4 \times 4$ control features of dimension 16 using gradient descent. We optimize with a learning rate of 1e-2 for 500 steps. 

\subsection{Temporal predictions}
\begin{figure}[!t]
    \centering
    \includegraphics[width=\linewidth]{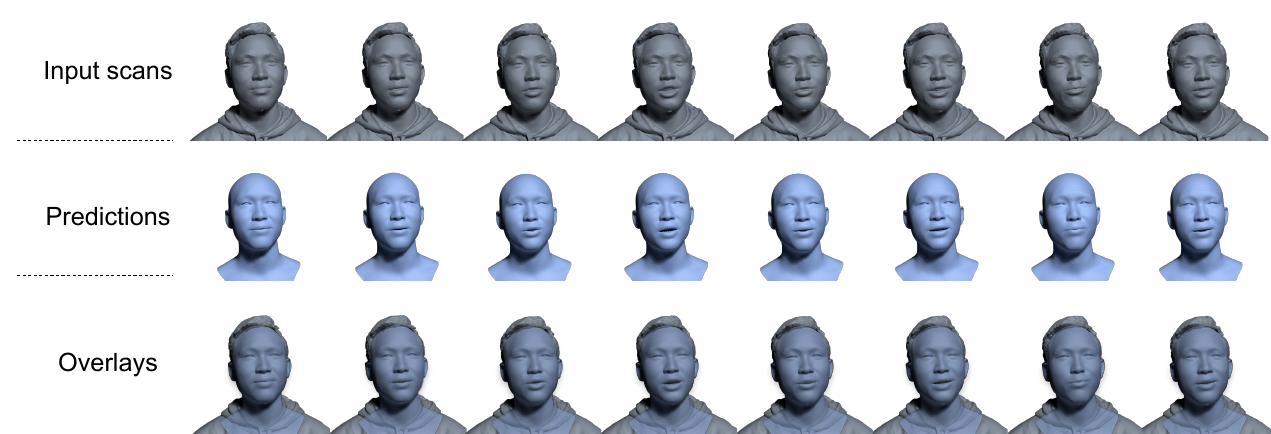}
    \caption{The \modelname scan encoder produces temporally smooth results for input scan sequences and can be used for performance registration. In this figure, we show the input scan, the prediction of our model and an overlay of the prediction on the input scan for a sequence of scans reconstructed from a facial performance of a subject. Note that the scans were processed independently for each frame.}
    \label{fig:scanseq}
    
    \vspace{2em}

    \includegraphics[width=\linewidth]{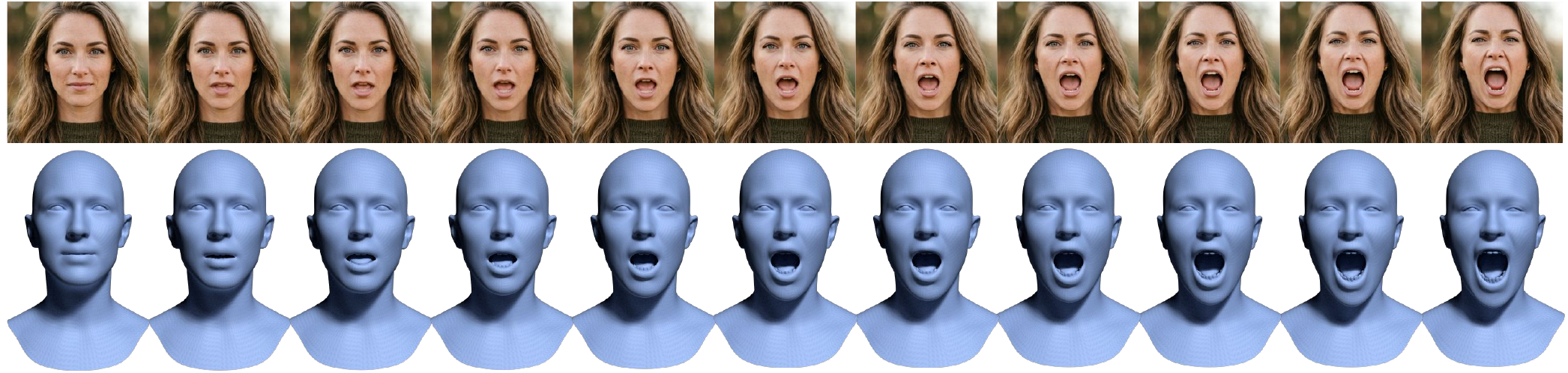}
    \caption{The \modelname image encoder can process frames from a video sequence to reconstruct facial performances in the wild. In this example, we show the captured images in the first row and the corresponding predictions from our model for each image in the second row. Our model produces a faithful reconstruction of the input facial performance.}
    \label{fig:imgseq}

    \vspace{28em}
    
\end{figure}

Our model produces temporally smooth predictions when applied to scan and image sequences. In \cref{fig:scanseq} and \cref{fig:imgseq} we show predictions from our model for sequential scan and image data respectively. Kindly have a look at our supplemental video for more results.

\subsection{Qualitative results}
\begin{figure*}
    \centering
    \includegraphics[width=\linewidth]{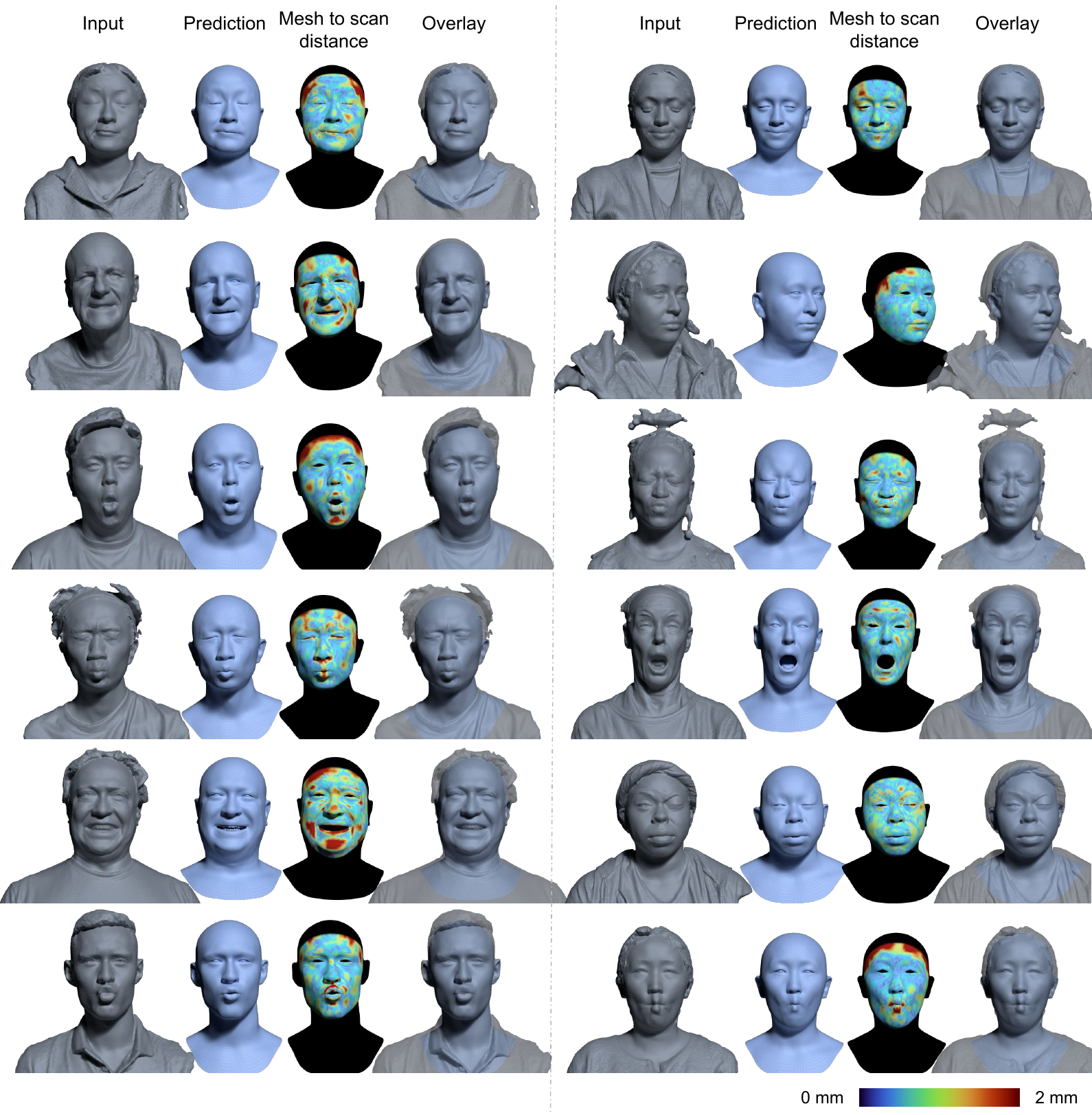}
    \caption{Additional qualitative results from our \modelname-Large $16 \times 16 \times 16$ model for scan registration. For each input scan, we visualize the prediction from the model and the point to scan distance (mm). Our method can handle input scans in varying topologies acquired from a diverse collection of subjects in various expressions and accessories.}
    \label{fig:scanadditional}
\end{figure*}

\begin{figure*}
    \centering
    \includegraphics[width=0.9\linewidth]{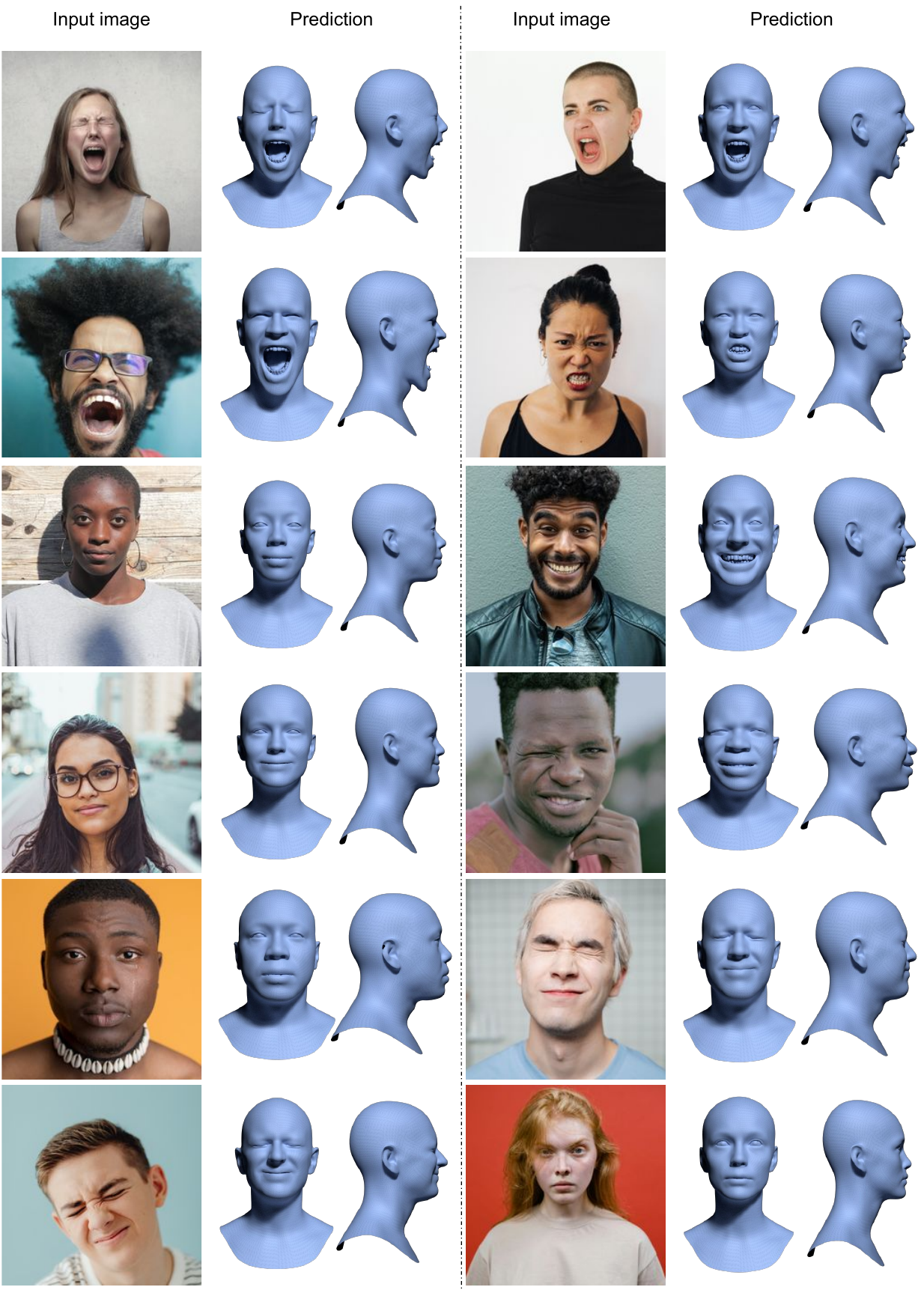}
    \caption{Additional qualitative results from our \modelname-Large $8 \times 8 \times 8$ model for face reconstruction from monocular images. For each input image, we visualize the predicted from the front and the side.}
    \label{fig:imageadditional}
\end{figure*}

Finally we provide additional qualitative results for our scan registration and image-based face reconstruction models in \cref{fig:scanadditional} and \cref{fig:imageadditional} respectively. These results highlight \modelname's ability to represent diverse face shapes and expressions while also being a continuous and localized shape representation.

\end{document}